\documentclass[10pt,twocolumn,letterpaper]{article}

\usepackage[pagenumbers]{cvpr} 

\usepackage{amsmath,amsfonts,bm}

\def\eqref#1{equation~\ref{#1}}

\def\1{\bm{1}}

\DeclareMathAlphabet{\mathsfit}{\encodingdefault}{\sfdefault}{m}{sl}
\SetMathAlphabet{\mathsfit}{bold}{\encodingdefault}{\sfdefault}{bx}{n}

\usepackage[utf8]{inputenc}
\usepackage[T1]{fontenc}
\usepackage{url}
\usepackage{booktabs}
\usepackage{amsfonts}
\usepackage{nicefrac}
\usepackage{microtype}
\usepackage{xcolor}
\usepackage{graphicx}
\usepackage{amsmath}
\usepackage{xspace}
\usepackage{enumitem}
\usepackage{subcaption}
\usepackage{epigraph}
\usepackage{changepage}
\usepackage{multirow}
\usepackage{colortbl}

\newcommand{\model}{RVM\xspace}

\definecolor{tabbestcolor}{rgb}{0.785, 0.851, 0.969}

\definecolor{grayoutcolor}{rgb}{0.785, 0.785, 0.785}

\newlength\savewidth\newcommand\shline{\noalign{\global\savewidth\arrayrulewidth
  \global\arrayrulewidth 1pt}\hline\noalign{\global\arrayrulewidth\savewidth}}
\newcommand{\tablestyle}[2]{\setlength{\tabcolsep}{#1}\renewcommand{\arraystretch}{#2}\centering\footnotesize}
\definecolor{baselinecolor}{HTML}{d6eaf8}
\newcommand{\baseline}[1]{\cellcolor{baselinecolor}{#1}}
\newcommand{\baselinerow}[1]{\rowcolor{baselinecolor}#1}

\newcolumntype{x}[1]{>{\centering\arraybackslash}p{#1pt}}
\newcolumntype{y}[1]{>{\raggedright\arraybackslash}p{#1pt}}
\newcolumntype{z}[1]{>{\raggedleft\arraybackslash}p{#1pt}}

\definecolor{highlight}{rgb}{0.87, 0.72, 0.53}
\newcolumntype{a}{>{\columncolor{highlight}}c}

\usepackage{listings}
\usepackage{xcolor}

\definecolor{codegreen}{rgb}{0,0.6,0}
\definecolor{codegray}{rgb}{0.5,0.5,0.5}
\definecolor{codepurple}{rgb}{0.58,0,0.82}

\definecolor{cvprblue}{rgb}{0.21,0.49,0.74}
\usepackage[pagebackref,breaklinks,colorlinks,allcolors=cvprblue]{hyperref}

\def\paperID{*} 
\def\confName{CVPR}
\def\confYear{2026}

\title{Recurrent Video Masked Autoencoders}

\author{Daniel Zoran, Nikhil Parthasarathy, Yi Yang, Drew A Hudson, Jo{\~a}o Carreira, Andrew Zisserman\\
{\small Google DeepMind}\\
}

\begin{document}
\twocolumn[{
  \maketitle
  \vspace{-20pt}
  \centering
  \href{https://rvm-paper.github.io}{https://rvm-paper.github.io}
  \vspace{10pt}
}]

\begin{abstract}

We present \textbf{R}ecurrent \textbf{V}ideo \textbf{M}asked-Autoencoders (\model): a novel approach to video representation learning that leverages recurrent computation to model the temporal structure of video data. \model couples an asymmetric masking objective with a transformer-based recurrent neural network to aggregate information over time, training solely on a simple pixel reconstruction loss. This design yields a highly efficient "generalist" encoder: \model achieves competitive performance with state-of-the-art video models (e.g. VideoMAE, V-JEPA) on video-level tasks like action classification, and point and object tracking, while matching or exceeding the performance of image models (e.g. DINOv2) on tasks that require strong geometric and dense spatial features. Notably, RVM achieves strong performance in the small-model regime without requiring knowledge distillation, exhibiting up to $30\times$ greater parameter efficiency than competing video masked autoencoders. Finally, we demonstrate that \model's recurrent nature allows for stable feature propagation over long temporal horizons with linear computational cost, overcoming some of the limitations of standard spatio-temporal attention-based video models. Ablation studies further highlight the factors driving the model's success, with qualitative results showing that RVM learns rich representations of scene semantics, structure, and motion.

\end{abstract}    
\section{Introduction}
\label{sec:intro}

\begin{figure}[ht]
\centering
\hspace{-0.07\textwidth}
\includegraphics[width=0.49\textwidth]{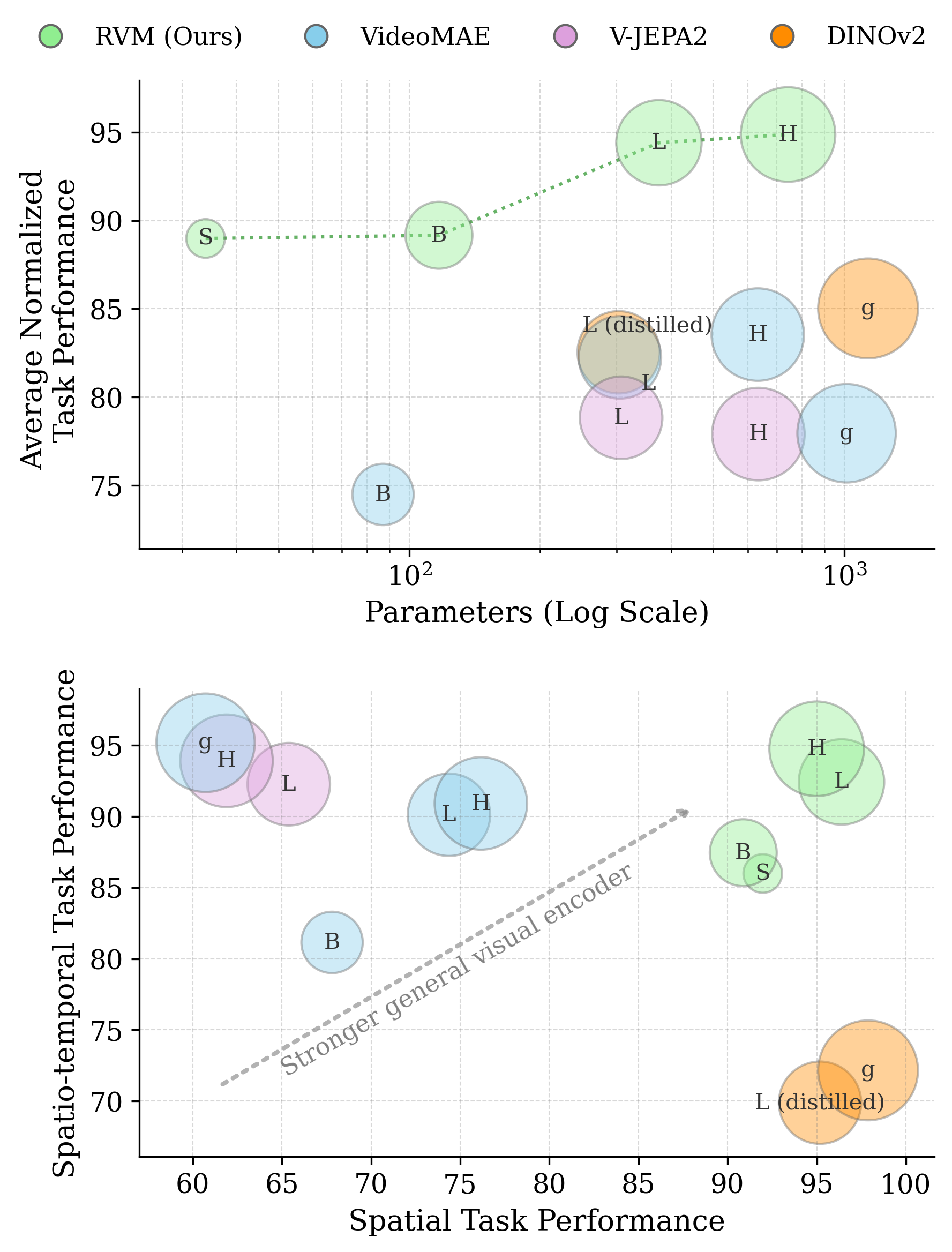}
\caption{Normalized task performance is calculated for each task (relative to the best model) and averaged across tasks. \textbf{Top:} Across a wide range of visual tasks that require strong spatiotemporal features (video) and dense spatial features (image), RVM models set a pareto frontier that outperforms other strong video and image encoders. Spatio-temporal tasks cover: Something Somethingv2, Kinetics, Waymo object tracking, Perception Test TAP; Spatial tasks cover: ScanNet depth and nearest-neighbor correspondence tasks (Davis segmentation, JHMDB, Video Instance Parsing). \textbf{Bottom:} RVM models bridge the gap between strong spatial task models (e.g.\ DINOv2) and strong video task encoders, achieving the best of both worlds. Circle sizes are proportional to model size.} 
\label{fig:splash}
\vspace{-1em}
\end{figure}

It has long been hypothesized that biological systems learn visual representations by predicting the spatio-temporal evolution of the world \citep{barlow1961possible, rao1999predictive, palmer2015predictive, singer2018sensory}. Indeed, even limited motion cues are sufficient to drive children’s ability to robustly perceive and segment objects \citep{spelke1990principles}. Recent advances in self-supervised learning (SSL) have revived the hope that artificial vision systems might also acquire such predictive world models purely from large-scale unlabeled video \citep{tong2022videomae, bardes2023vjepa}.

Among the most successful approaches are masked autoencoders (MAEs), which learn by reconstructing randomly masked portions of images or videos \citep{he2022masked, tong2022videomae, feichtenhofer2022masked, wang2023videomae, carreira2024scaling}, and Joint Embedding Predictive Architectures (JEPAs) \citep{bardes2023vjepa, vjepa2}, which predict future states in latent space while avoiding collapse via architectural or training heuristics \citep{ajepa, mcjepa, sjepa}. Latent-space prediction has been argued to encourage learning task-relevant representations by discarding nuisance factors \citep{grill2020bootstrap}.

For video, both VideoMAE \citep{tong2022videomae, wang2023videomae} and V-JEPA \citep{bardes2023vjepa, vjepa2} rely on early-fusion spatio-temporal encoders (with spatio-temporal attention throughout the network) and random masking across entire clips. These designs treat time as uniform and symmetric, both in masking and in attention, neglecting the causal and directional nature of temporal dynamics. As a result, they are less amenable to online or streaming applications such as robotics. Moreover, their chunked offline architectures limit inference to short clips, preventing consistent representation learning over longer horizons.

Conversely, image-based models such as DINO \citep{radford2021learningtransferablevisualmodels, oquab2023dinov2} excel at learning semantic representations and provide stable features when unrolled over multiple frames, but as they are still \textit{image} models they are unable to capture motion information in their features. SiamMAE \citep{gupta2023siamese} partially addresses these limitations by training image encoders on natural video data, incorporating temporal asymmetry: it conditions on an unmasked “past” (\emph{source}) frame to reconstruct a heavily masked “future” (\emph{target}) frame via a cross-attention decoder. This asymmetric setup provides a strong inductive bias for learning correspondences. Nevertheless, SiamMAE still trains an \emph{image} encoder, and thus it cannot ultimately capture spatio-temporal dependencies needed for true video-centric tasks.

In this work, we propose \textbf{R}ecurrent \textbf{V}ideo \textbf{M}asked Autoencoders (RVM), a family of general visual encoders that (in the spirit of SiamMAE), explicitly model the asymmetry of time through both masking and architecture. The RVM architecture processes videos sequentially by aggregating frame-level representations via a recurrent module. Training only with a pixel reconstruction loss on large natural video data, RVM learns strong representations that set a new pareto frontier in parameter efficiency (Figure \ref{fig:splash}, top) when evaluated across a wide-range of visual tasks. While SoTA models tend to specialize, RVM is \textit{uniquely general, achieving strong average performance across spatial and video (spatio-temporal) tasks} (Figure \ref{fig:splash}, bottom). Furthermore, in the small model regime, RVM performs strikingly well \textit{without requiring any form of model distillation}. Finally, owing to its recurrent design, RVM features \textit{show emergent feature stability at long time horizons}, while being able to be unrolled over such sequences with linear compute and memory.

\section{Related Work}
\label{sec:relwork}
\textbf{Self-Supervised Video Models.} Over the recent years, Self-Supervised Learning (SSL) \citep{he2022masked, chen2020simple, caron2021emerging, jaiswal2020survey} has became a leading paradigm for deriving powerful representations from unlabled visual data. In the case of videos, diverse learning methods have been proposed that harness the rich spatio-temporal nature inherent to the domain. Earlier approaches focused on pretext tasks \citep{doersch2015unsupervised} that were designed to encourage learning temporal coherence and dynamics, by predicting frame order \citep{lee2017unsupervised, misra2016shuffle, xu2019self}, motion statistics \citep{wang2015unsupervised, pathak2017learning, agrawal2015learning} or playback speed \citep{benaim2020speednet, pickup2014seeing, yao2020video}. Other methods leveraged the multi-modal correspondence of video and audio, aiming to predict synchronization between the two \citep{chung2016out}. More recently, contrastive learning approaches \citep{pan2021videomoco, qian2021spatiotemporal, han2020self, li2022bridge} were developed to encourage consecutive frames' embeddings to stay closer in the latent space, while pushing frames from different videos apart \citep{xu2021rethinking}. Meanwhile, masked modeling approaches \citep{tong2022videomae, wang2022bevt, hong2022cogvideo} have proven both effective and robust in learning rich context-aware video representations, by reconstructing masked spatio-temporal patches from their surroundings.

\noindent \textbf{Masked Autoencoders.} Within the masked-modeling paradigm \citep{he2022masked, bao2021beit, xie2022simmim}, a range of works introduce architectural and objective-level extensions to accommodate multiple views or frames \citep{tong2022videomae, wang2022bevt, hong2022cogvideo}. A prominent direction involves the integration of Siamese networks \citep{bertinetto2016fully, chicco2021siamese} and Masked Image Modeling \citep{
he2022masked, bao2021beit, xie2022simmim}, with examples such as SiamMAE \citep{gupta2023siamese}, CropMAE \citep{eymael2024efficient} and CroCo \citep{weinzaepfel2022croco} that respectively reconstruct a heavily-masked frame, crop, or view of a 3D-scene, by conditioning on a second unmasked one. Likewise, Counterfactual World Modeling (CWM) \citep{bear2023unifying} explore temporally-factored masking, where a fully-visible frame informs the prediction of its heavily-occluded subsequent. Guided Future Prediction \citep{carreira2024learning} innovates over standard masking, by replacing a few patches of an input frame with respective patches from a future one, to guide its reconstruction. Alternatively, MotionMAE \citep{yang2024motionmae} directly enriches the standard masking objective with the prediction of temporal difference between successive frames, so to encourage modeling of motion and dynamics. 

\noindent \textbf{Recurrent Video Models.} Compared to the methods discussed above, our self-supervised video model stands out as it processes videos recurrently, so to explicitly model their temporal dynamics. It links to prior works about recurrent video architectures \citep{donahue2015long, yue2015beyond}. One example is the Recurrent Vision Transformer (RViT) model \citep{yang2022recurring}, which forms an aggregated representation of a video by processing it iteratively with attention-based gating. Another notable instance is the Recurrent Convolutional Neural Network (RCNN) \citep{liang2015recurrent}, which embeds recurrent connections directly into its convolutional layers, enabling the model to learn spatio-temporal features in a unified manner. By adopting a recurrent processing scheme, these models can assimilate the progression and directionality of time, and fully capture the long-range dependencies across frames and the temporal dynamics of videos. Recently, State Space Models (SSMs) such as VideoMamba \citep{li2024videomamba} and VideoMambaPro \citep{lu2024videomambapro} have also been explored for efficient video understanding. However, unlike our approach, these methods typically rely on non-causal, bidirectional processing to achieve competitive performance and process videos as a flat sequence of tokens, discarding the spatial information.
\begin{figure*}[ht]
    \centering
    \includegraphics[width=0.8\textwidth]{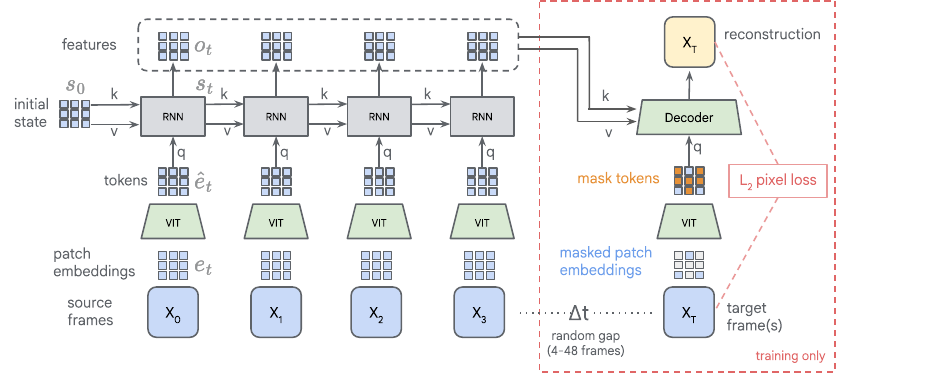}
    \caption{\textbf{\model overview.} The model encodes \textit{source} frames from an input video sequentially. Each frame is independently encoded using a vision transformer and the output tokens are aggregated using a transformer-based RNN to produce a sequence of features. See text for full details.
    During training, a \textit{target} frame is sampled from a random time gap in the future, masked and encoded using the same ViT encoder. The model is trained to reconstruct the masked target frame using a cross attention decoder, minimizing the $L_2$ loss between reconstruction and target.}
    \label{fig:model}
    \vspace{-1em}
\end{figure*}

\section{Model}
\label{sec:model}

\model is a recurrent model that encodes frames $X_t$ sequentially to produce a set of features for each frame. Figure \ref{fig:model} provides an overview of the general architecture. Each input frame is patchified and encoded using a ViT. The resulting tokens are fed into a Recurrent Neural Network (RNN) core, which carries a state from the previous time-step and integrates tokens from the current frame to produce an updated state. This new state serves as the feature representation for the current time-step. By processing sequentially, \model is able to ingest, discard, and refine information incrementally as it becomes available.

During training, the also model receives a \textit{target} frame $X_T$ sampled from a (potentially distant) future. We sample the target frame with a random time gap $\Delta t$ of between 4 and 48 frames from the last source frame. Depending on the exact data used this corresponds to a time gap of 0.15 to 10 seconds. This target frame is then heavily masked and encoded using the same encoder as the input (\textit{source}) frames. The training objective is to reconstruct the target frame using information from the source frames. This is achieved using a cross-attention based decoder \citep{gupta2023siamese} by minimizing the $L_2$ loss between the reconstruction and the target.

\subsection{Modules}
\paragraph{Tokenization \& Masking}
Each frame $X_t\in \mathbb{R}^{H\times W\times 3}$ (both source and target) is divided into non-overlapping patches of size $P \times P$. These patches are embedded via a learnable linear projection, resulting in a feature map of size $h \times w \times D$ (where $h=H/P$ and $w=W/P$). These embeddings are then flattened into a sequence of $N=hw$ tokens of size $D$. This tokenization process is applied independently to each frame, with weights shared across all source and target frames. Fourier positional encodings are subsequently added to the tokens.

For the \textit{target} frame, tokens are randomly masked with a ratio $m$ (defaulting to $m=0.95$). A learnable \texttt{[CLS]} token is concatenated to the token sequences of both source and target frames. This yields a sequence of $K$ source token sets, denoted as $e^{\texttt{S}}_1, \dots, e^{\texttt{S}}_K$ where $e^\texttt{S}_t \in \mathbb{R}^{(N+1) \times D}$, and a single set of unmasked target tokens $e^\texttt{T} \in \mathbb{R}^{(M+1) \times D}$, where $M = \lfloor (1-m)N \rfloor$.
\vspace{-1em}
\paragraph{Encoder}
We employ a ViT encoder \citep{dosovitskiy2020image} to process the tokens of each frame independently. Specifically, we utilize standard ViT blocks with pre-normalization and without dropout. Following SiamMAE \citep{gupta2023siamese}, the encoder weights are shared across all frames, both source and target. We denote the resulting encoded outputs as $\hat{e}_t^S$ for the source frames ($t=1 \dots K$) and $\hat{e}^T$ for the target frame.

\vspace{-1em}

\paragraph{Recurrent Core}
The encoded outputs from the source frames are fed into a \textit{recurrent neural network} (RNN) core, formally defined as $o_t, s_t = R(x_t, s_{t-1})$. Here, $x_t$ represents the input at the current time step, $s_{t-1}$ denotes the state from the previous time step, while $o_t$ and $s_t$ represent the output and updated state for the current time step, respectively. We unroll the RNN sequentially over the source frames, producing a sequence of outputs $o_t \in \mathbb{R}^{(N+1)\times D}$ and states $s_t \in \mathbb{R}^{(N+1)\times D}$ for $t=1 \dots K$. The initial state $s_0$ is set to zero. This recurrent mechanism enables the model to aggregate information over time, constructing a temporally-aware representation. We utilize the outputs $o_t$ as features for downstream tasks. The specific architectural details of the RNN are discussed in Section \ref{sec:gru}.
\vspace{-1em}
\paragraph{Decoder}
During training, our objective is to reconstruct the target frame $X^T$ from its masked tokens $\hat{e}^T$, conditioned on the source frame features $o_t$. We employ a decoder with both cross- and self-attention mechanisms, similar to \citep{gupta2023siamese}. Both target and source frame features are first embedded via a linear layer. Following the standard MAE approach \citep{he2022masked}, we place the unmasked target tokens into their original grid positions, fill the masked locations with a learnable \texttt{[MASK]} token, and add Fourier positional embeddings. This sequence serves as the input to the decoder.

Each decoder block consists of three sequential components: (1) cross-attention, utilizing target tokens as queries and source tokens (concatenated along the token axis) as keys and values; (2) a feed-forward MLP; and (3) self-attention. All components utilize residual connections and pre-normalization (LayerNorm \citep{layernorm}). Finally, the decoder output is projected to the original patch dimension and reshaped to reconstruct the target frame.
\vspace{-1em}
\paragraph{Loss}
We use a simple $L_2$ loss over the entire reconstructed and target image pixels, with no patch level normalization.

\subsection{The Rise of GRU}
\label{sec:gru}

To effectively aggregate and integrate information over time, our model requires a module capable of maintaining a \textit{state} across time steps. Ideally, this mechanism should retain critical information, discard irrelevant data, and assimilate new inputs as they arrive. Furthermore, we seek to leverage the efficacy of Transformers to facilitate spatiotemporal interactions between tokens. To address these needs, we propose a hybrid architecture combining a Transformer with a Gated Recurrent Unit (GRU).

This RNN core utilizes a combination of cross- and self-attention to integrate information. Specifically, the encoder outputs $\hat{e}_t$ for the current time step serve as queries, while the keys and values are derived from the previous state $s_{t-1}$. To manage this information flow, we adopt the gating mechanism of the standard GRU \citep{cho2014properties}. The \textit{reset gate} $r_t$ modulates the previous state before it is passed to the attention block, while the \textit{update gate} $u_t$ determines the balance between the previous state and the new attention output. The module is governed by the following equations:
\begin{align}
\nonumber
    u_t &= \sigma\left(W^u_e \hat{e}_t + W^u_s s_{t-1}\right) \\
\nonumber
    r_t &= \sigma\left(W^r_e \hat{e}_t + W^r_s s_{t-1}\right) \\
\nonumber
    \hat{h}_t &= \texttt{Tx}(q=\hat{e}_t, kv=r_t\odot s_{t-1}) \\
\nonumber
    s_t &= (1-u_t)\odot s_{t-1} + u_t \odot \hat{h}_t \\
\nonumber
    o_t &= s_t
\end{align}
Here, $\sigma$ denotes the sigmoid function, and $\texttt{Tx}$ represents a multi-layer Transformer block utilizing both cross- and self-attention. The weight matrices $W$ are applied to the feature dimension and shared across all tokens. The state $s_0$ is initialized to zero. Pseudo-code is provided in the Supplementary Material.
\section{Experiments}

\subsection{Training}
We train the model on a large dataset of a mixture of publicly available web videos. We base the mixture off of the mixture used in \cite{assran2025v}, containing sampled video clips from HowTo100M \citep{miech2019howto100m}, Kinetics700 \citep{carreira2019short}, SSV2 \citep{goyal2017something}, YTBB \citep{real2017youtube}, and YT8M \citep{abu2016youtube}. The full dataset contains approximately 8.4M video clips (For more details see Supplementary Material). During training we randomly sample sub-clips, applying random flipping and random resized crop augmentation. The final frames are resized to $256\times256$ resolution. We train several model sizes, scaling the encoder and RNN core accordingly, while following standard masked autoencoder (MAE) practice by keeping the decoder size fixed across experiments. See Supplementary Material for architectural details (number of layers, hidden dimensions, etc.). All models are trained \textit{from scratch} and no distillation procedure is used.

Unless otherwise stated, models are trained for 1M steps (250k steps for ablations) with a global batch size of 2048, corresponding to about 2B training examples in total. We highlight that 
the \model architecture and objective seems to enable training for very long schedules with steady performance increase in all downstream tasks. 

Each training example consists of a 64-frame video clip sampled randomly from the dataset mixture. From each clip, we sample 4 consecutive source frames that are processed by the recurrent encoder. We reconstruct 4 target frames, that are sampled independently and uniformly between 4 and 48 frames after the last source frame. Training is distributed across 256 TPU-v6 cores with per-core memory of 32GB. We use \texttt{bfloat16} precision for all forward and backward passes, while upcasting the loss and softmax computations to \texttt{float32}; model weights are stored in  \texttt{float32} for stability. To fit large models within device memory, we employ FSDP-like parameter and optimizer sharding. Optimization uses AdamW \citep{loshchilov2017decoupled} with a cosine decay learning rate schedule and warm-up phase. Full hyperparameter settings are summarized in the Supplementary Material.

\subsection{Quantitative Results}
\textbf{Baselines.} We compare RVM against a set of strong image and video model baselines: 
\begin{itemize}
\item \textbf{Image models}: We compare to DINOv2 \citep{oquab2023dinov2} as the main baseline for strong spatial task performance. At small model scales, we also include SiamMAE \citep{gupta2023siamese} as it is a strong efficient model for dense correspondence tasks such as video segmentation and human keypoint tracking.

\item \textbf{Video models}: We evaluate variants of VideoMAE \citep{tong2022videomae},

V-JEPA2 \citep{vjepa2}, and 4DS \citep{carreira2024scaling}. These models are designed for large-scale video pretraining and largely represent the current frontier in video self supervised learning. 
\end{itemize}
\begin{figure}
    \centering
    \begin{subfigure}[t]{0.15\textwidth}
        \centering
        \includegraphics[width=\textwidth, height=\textwidth]{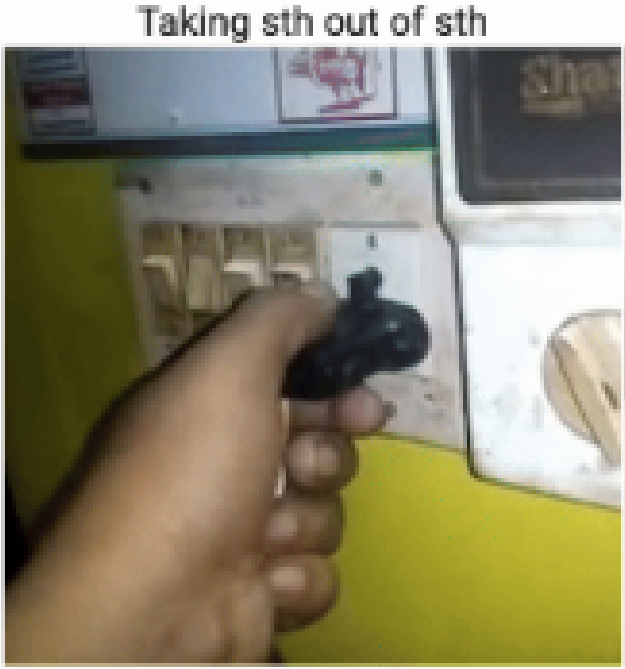}
        \captionsetup{font=tiny}
        \caption{Action recognition on SSv2 \citep{ssv2} and Kinetics \citep{kinetics}.}
        \label{fig:task_a}
    \end{subfigure}
    \hspace{0.001\textwidth} 
    \begin{subfigure}[t]{0.15\textwidth}
        \centering
        \includegraphics[width=\textwidth, height=\textwidth]{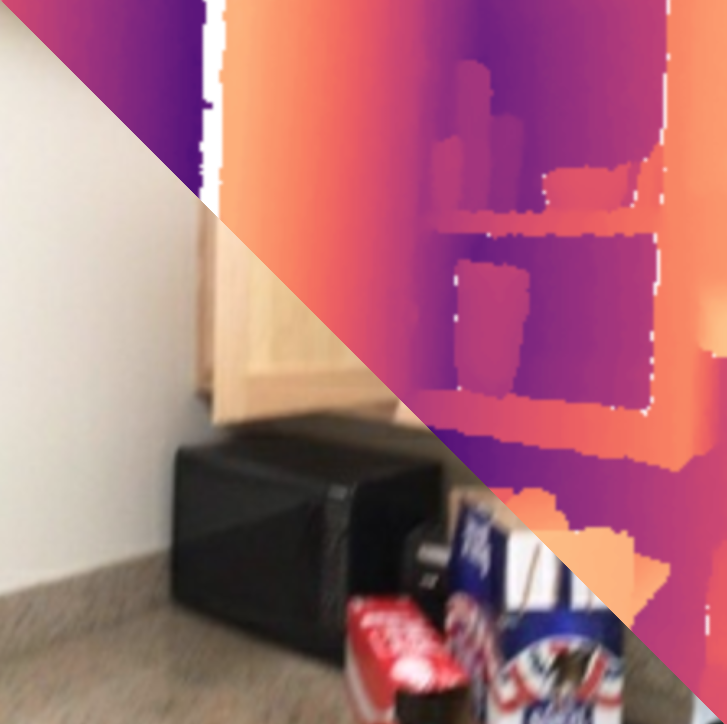}
        \captionsetup{font=tiny}
        \caption{Depth estimation on ScanNet \citep{dai2017scannet}.}
        \label{fig:task_b}
    \end{subfigure}
    \hspace{0.001\textwidth} 
    \begin{subfigure}[t]{0.15\textwidth}
        \centering
        \includegraphics[width=\textwidth, height=\textwidth]{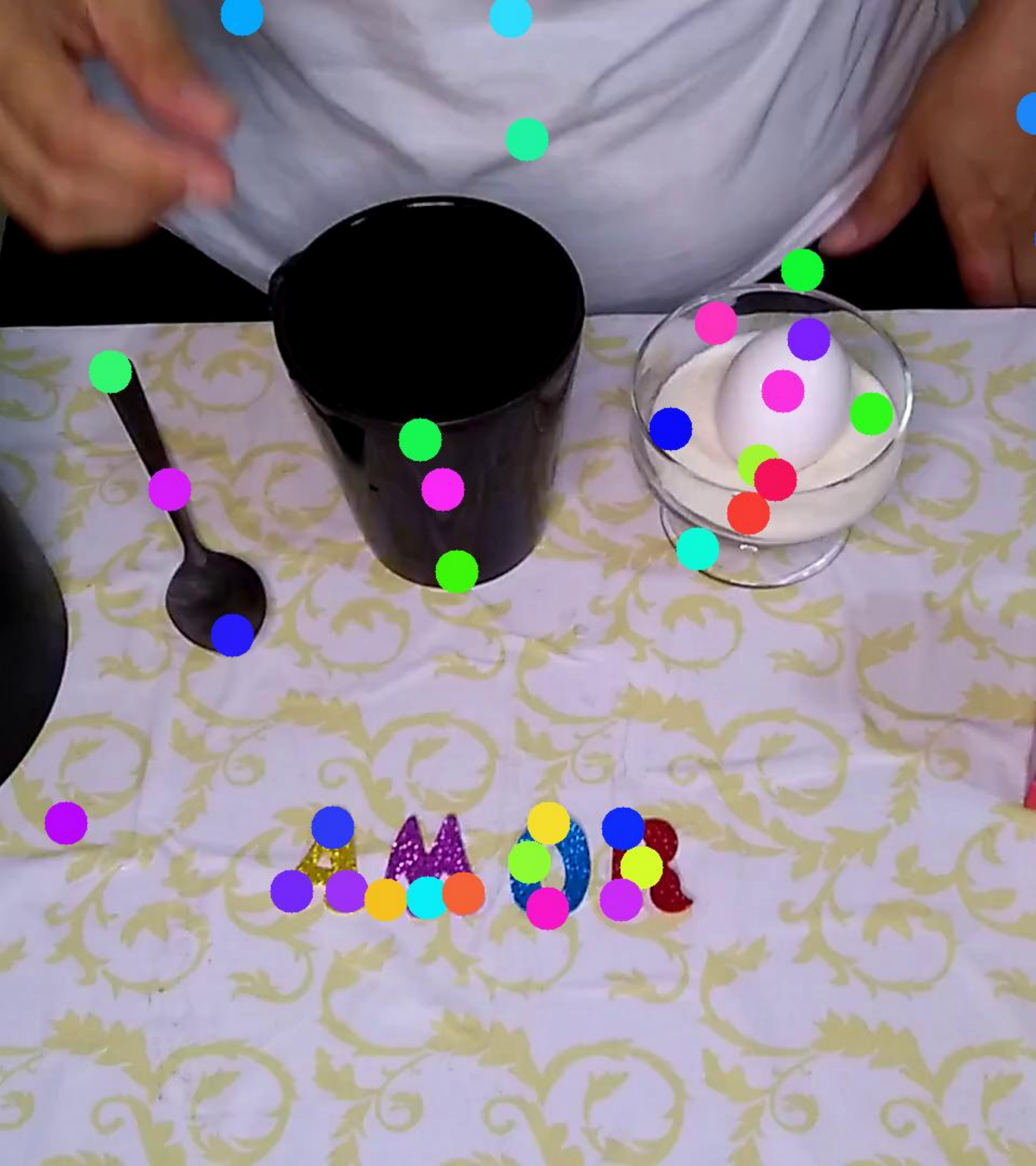}
        \captionsetup{font=tiny}
        \caption{Point tracking on Perception Test \citep{patraucean2023perception}.}
        \label{fig:task_c}
    \end{subfigure}
    \hspace{0.001\textwidth} 
    \begin{subfigure}[t]{0.15\textwidth}
        \centering
        \includegraphics[width=\textwidth, height=\textwidth]{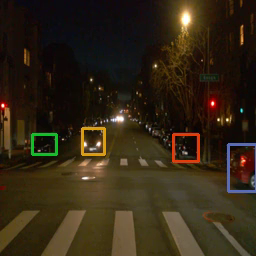}
        \captionsetup{font=tiny}
        \caption{Object tracking on Waymo Open \citep{sun2020waymo}.}
        \label{fig:task_d}
    \end{subfigure}
    \hspace{0.001\textwidth} 
    \begin{subfigure}[t]{0.15\textwidth}
        \centering
        \includegraphics[width=\textwidth, height=\textwidth]{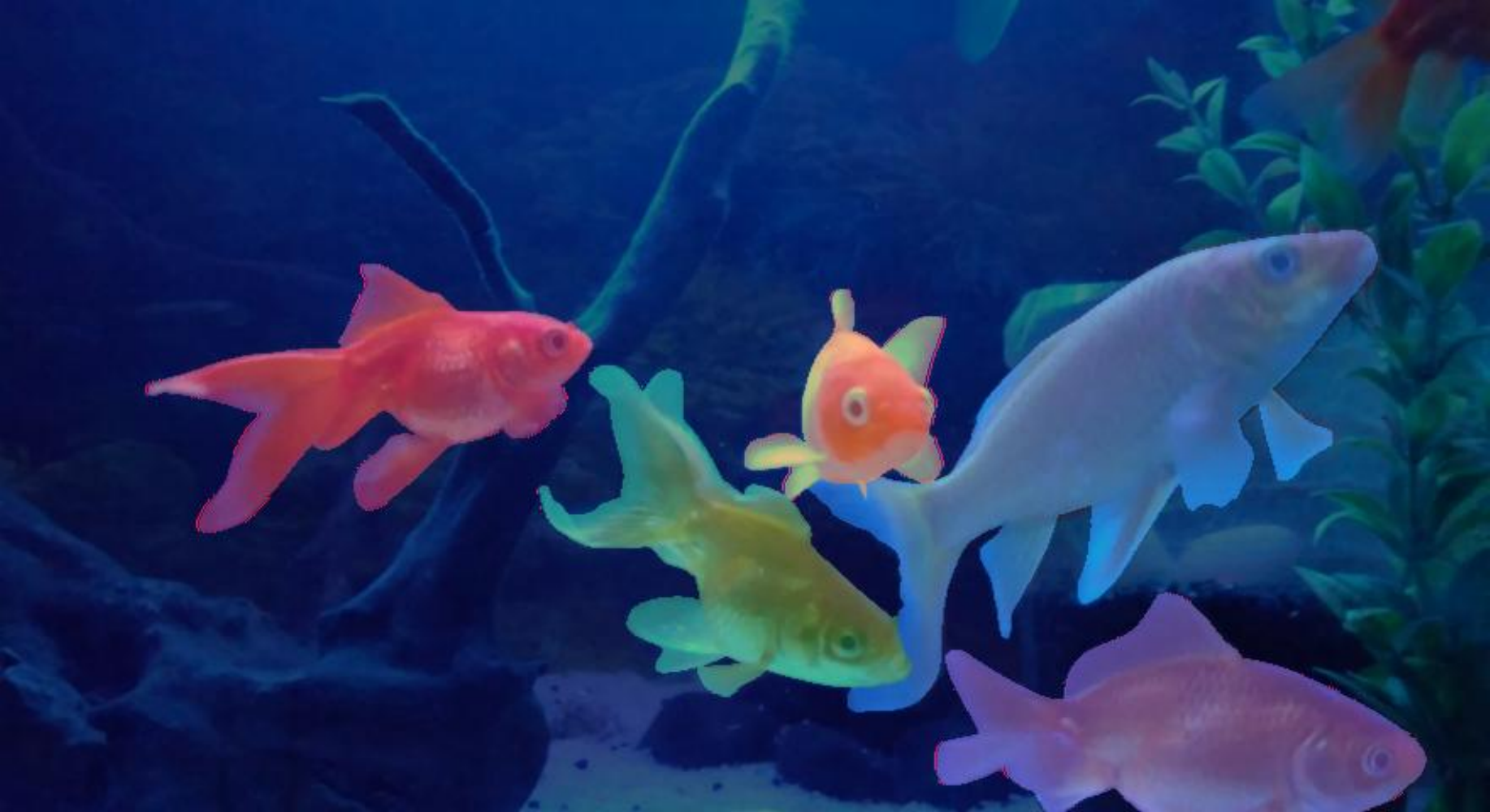}
        \captionsetup{font=tiny}
        \caption{Segmentation tracking on DAVIS \citep{pont20172017} and VIP \citep{xu2018video}}
        \label{fig:task_e}
    \end{subfigure}
    \hspace{0.001\textwidth} 
    \begin{subfigure}[t]{0.15\textwidth}
        \centering
        \includegraphics[width=\textwidth, height=\textwidth]{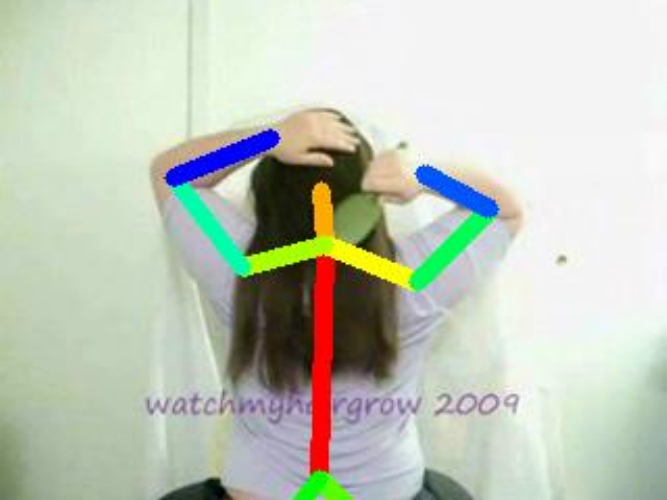}
        \captionsetup{font=tiny}
        \caption{Keypoint tracking on JHMDB \citep{jhuang2013towards}.}
        \label{fig:task_f}
    \end{subfigure}    
   
    \caption{\textbf{Evaluation suite.} Individual frames and annotations from some of the evaluation tasks in this paper, covering semantic, geometry and motion perception.}
    \label{fig:tasks}
    \vspace{-0.5em}
\end{figure} 
\textbf{Evaluation Suite.} We evaluate \model on a diverse set of benchmarks (Figure~\ref{fig:tasks} that we group into two primary categories: 
\begin{itemize}
\item \textbf{Spatio-temporal tasks}: where performance generally improves with better modeling of both semantic/spatial and motion features. This set contains action recognition (SSv2 and Kinetics-700) \citep{ssv2, kinetics}, Waymo open object tracking \citep{sun2020waymo}, and Perception Test point tracking \citep{patraucean2023perception}. 
\item \textbf{Spatial tasks}: where strong, dens features extracted independently at the frame-level are generally sufficient. This set contains, Scannet depth estimation \citep{dai2017scannet}, DAVIS-2017 segmentation (spatial correspondence across frames) \citep{pont20172017}, JHMDB human keypoint tracking \citep{jhuang2013towards}, and VIP human part tracking \citep{xu2018video}. 
\end{itemize} \vspace{0.5em}
\textbf{Evaluation Protocol.} From a functional perspective, the tasks above can be also categorized into ``readout tasks'', which train readout heads on top of frozen encoder representations, and ``nearest-neighbor'' (zero-shot) tasks that use pixel-level semantic label-propagation. For readout tasks, on top of the \textit{frozen} pre-trained model we train an attentive readout head that follows the exact protocol described in recent literature \citep{carreira2024scaling}.  Read-out heads for all models, including external ones, are trained using the same setup, with publicly available checkpoints. Nearest-neighbor tasks perform various forms of label propagation that follow the original protocols from each evaluation dataset. For more details on these benchmarks and the evaluation protocol, see the Supplementary Material.

\newcommand{\tbest}[1]{\textbf{\textit{#1}}} 

\begin{table*}
\centering
\caption{\textbf{RVM learns a general visual representation that succeeds at both video-centric tasks that require spatio-temporal representations as well as tasks that primarily require strong dense geometric and spatial features.} While RVM does not outperform all baselines on every benchmark, we see that it provides the strongest general representation across all tasks, indicated by the Avg. normalized accuracy. We compute this by averaging the scores for each model across tasks after normalizing each column by the best model performance. }
\label{tab:lh_models}
\resizebox{\textwidth}{!}{
\begin{tabular}{lcccccccccc}
\toprule
& & \multicolumn{4}{c}{Spatio-temporal tasks} & \multicolumn{4}{c}{Spatial Tasks} & \\
\cmidrule(r){3-6} \cmidrule(r){7-10}
Model & Size(M) & SSv2 & Kinetics & Waymo & PT & ScanNet & DAVIS & JHMDB & VIP & \textbf{Average} \\
& & Acc. ($\uparrow$\%) & Acc. ($\uparrow$\%) & mIoU ($\uparrow$) & AJ ($\uparrow$) & AbsRel ($\downarrow$) & J\&F ($\uparrow$\%) & PCK@0.1 ($\uparrow$\%) & mIoU ($\uparrow$) & \textbf{Normalized ($\uparrow$)} \\
\midrule

4DS-L & 310 & 57.6 & 45.2 & \textbf{75.9} & \textbf{81.5} & 1.23 & -- & -- & -- & -- \\
VideoMAE-L & 305 & 62.7 & 52.5 & 74.9 & 78.3 & 1.10 & 54.3 & 45.1 & 18.9 & 82.2 \\
V-JEPA2-L & 307 & \textbf{67.5} & \textbf{59.2} & 73.1 & 73.7 & 1.08 & 47.5 & 33.3 & 17.2 & 78.8 \\

\baselinerow \model-L & 375 & 66.7 & 57.3 & 73.2 & 77.3 & \textbf{0.91} & \textbf{66.0} & \textbf{48.4} & \textbf{38.0} & \textbf{94.4} \\

\textcolor{gray}{DINOv2-L (distilled)} & \textcolor{gray}{303} & \textcolor{gray}{52.2} & \textcolor{gray}{\textbf{63.5}} & \textcolor{gray}{51.7} & \textcolor{gray}{36.6} & \textcolor{gray}{1.02} & \textcolor{gray}{61.7} & \textcolor{gray}{\textbf{51.4}} & \textcolor{gray}{\textbf{40.6}} & 82.5 \\
\midrule
4DS-H & 639 & 60.0 & 47.5 & \textbf{76.1} & \textbf{81.8} & 1.14 & -- & -- & -- & -- \\
VideoMAE-H & 633 & 64.2 & 54.5 & 74.6 & 77.1 & 1.04 & 54.4 & 43.9 & 20.8 & 83.5 \\
V-JEPA2-H & 635 & \textbf{68.9} & 59.1 & 73.6 & 77.0 & 1.05 & 43.3 & 29.7 & 16.0 & 77.9 \\

\baselinerow \model-H & 743 & 68.7 & \textbf{60.0} & 74.2 & 78.3 & \textbf{0.89} & \textbf{65.6} & \textbf{45.6} & \textbf{37.3} & \textbf{94.9} \\
\textcolor{gray}{DINOv2-g} & \textcolor{gray}{1135} & \textcolor{gray}{54.8} & \textcolor{gray}{65.4} & \textcolor{gray}{50.6} & \textcolor{gray}{39.9} & \textcolor{gray}{0.91} & \textcolor{gray}{62.4} & \textcolor{gray}{\textbf{51.1}} & \textcolor{gray}{\textbf{40.5}} & \textcolor{gray}{85.0} \\
\textcolor{gray}{VideoMAEv2-g} & \textcolor{gray}{1013} & \textcolor{gray}{65.6} & \textcolor{gray}{\textbf{69.7}} & \textcolor{gray}{72.6} & \textcolor{gray}{73.7} & \textcolor{gray}{1.08} & \textcolor{gray}{41.7} & \textcolor{gray}{28.6} & \textcolor{gray}{16.9} & \textcolor{gray}{77.9} \\
\bottomrule
\end{tabular}}
\end{table*}

\begin{table*}
\centering
\caption{\textbf{RVM enables strong small model performance without distillation.}} 
\label{tab:sb_models}
\resizebox{\textwidth}{!}{
\begin{tabular}{lcccccccccc}
\toprule
& & \multicolumn{4}{c}{Spatio-temporal Tasks} & \multicolumn{4}{c}{Spatial Tasks} & \\
\cmidrule(r){3-6} \cmidrule(r){7-10}
Model & Size(M) & SSv2 & Kinetics & Waymo & PT & ScanNet & DAVIS & JHMDB & VIP & \textbf{Average} \\
& & Acc. ($\uparrow$\%) & Acc. ($\uparrow$\%) & mIoU ($\uparrow$) & AJ ($\uparrow$) & AbsRel ($\downarrow$) & J\&F ($\uparrow$\%) & PCK@0.1 ($\uparrow$\%) & mIoU ($\uparrow$) & \textbf{Normalized ($\uparrow$)}  \\
\midrule
SiamMAE-S & 27 & 40.0 & 41.2 & 55.0 & 65.0 & 1.63 & 62.0 & 47.0 & \textbf{37.3} & 80.8  \\
4DS-S & 24 & 39.9 & 28.1 & 69.6 & 75.9 & 2.05 & -- & -- & -- & --  \\
\baselinerow \model-S & 34 & \textbf{59.7} & \textbf{49.6} & \textbf{70.5} & \textbf{76.5} & \textbf{0.97} & \textbf{62.9} & \textbf{47.5} & 35.9 & \textbf{96.1} \\
\textcolor{gray}{DINOv2-S (distilled)} & \textcolor{gray}{21} & \textcolor{gray}{48.3} & \textcolor{gray}{\textbf{57.1}} & \textcolor{gray}{51.0} & \textcolor{gray}{33.4} & \textcolor{gray}{1.17} & \textcolor{gray}{62.4} & \textcolor{gray}{\textbf{48.4}} & \textcolor{gray}{\textbf{39.4}} & 84.4 \\
\textcolor{gray}{VideoMAE-B} & \textcolor{gray}{87} & \textcolor{gray}{52.3} & \textcolor{gray}{38.9} & \textcolor{gray}{\textbf{73.1}} & \textcolor{gray}{\textbf{79.2}} & \textcolor{gray}{1.50} & \textcolor{gray}{50.6} & \textcolor{gray}{44.7} & \textcolor{gray}{19.6} & 80.4  \\

\bottomrule
\end{tabular}}
\vspace{-0.5em}
\end{table*}

\subsubsection{\model learns strong generalist vision models}

Results for large-scale models (L/H) in Table \ref{tab:lh_models} reveal a clear dichotomy in baseline performance. As an image-encoder, DINOv2 performs well on spatial and semantic tasks but fails on intensive spatio-temporal tasks (e.g., 36.6/39.9 point tracking vs. video models achieving > 70). In contrast, native video-encoders (VideoMAE, V-JEPA2) achieve high scores on spatio-temporal benchmarks like SSv2, but trade this off with very poor results on spatial correspondence (e.g. < 20 mIoU on VIP vs. 40 mIoU for DINOv2) .

\model unifies these capabilities, achieving strong performance across both axes. To quantify this balance, we compute a ``Normalized avg.'' by averaging model scores normalized against the best performance per benchmark. Under this metric, \model-L and \model-H not only outperform their direct counterparts by more than 10\% but also surpass giant-scale models (DINOv2-g, VideoMAEv2-g) by a similar margin. While the baselines show high variance with task-specific failures, \model is the only architecture to avoid poor performance across the entire evaluation suite.

\subsubsection{\model learns strong small models without distilation}
Table \ref{tab:sb_models} highlights the architectural efficiency of our approach. A key finding is that \model{} continues to yield strong performance with very efficient models (ViT-S scale) \emph{without requiring additional knowledge distillation.}

While there are not many model classes that even attempt to train at this scale, \model-S outperforms competing baselines of similar size, achieving notable gains on SSv2 (+3.7\%) and Kinetics-400 (+21.5\% over 4DS-S) and outperforming the frame-based SiamMAE model on 3 out of 4 spatial tasks. \model-S even outperforms a VideoMAE-B model (4x larger) on 6 out of 8 evaluations. This stands in contrast to prior state-of-the-art methods like DINOv2, that rely on distillation from larger teacher models to ensure strong performance in the small-compute regime. While distillation is clearly effective (DINOV2-S (distilled) outperforms \model-S on 3 benchmarks), it remains the case that \model-S has the highest average normalized performance across all benchmarks.

In fact, as seen in Figure. \ref{fig:splash} (Top), because \model does not perform poorly on any one task, the average normalized accuracy of \model-S \textit{even when aggregating across all model scales} still outperforms both VideoMAEv2-g and DINOv2-g (\emph{which are 30x larger!}).

\subsubsection{Long-term feature consistency}

\begin{figure}[ht]
     \centering
     \includegraphics[width=0.9\columnwidth]{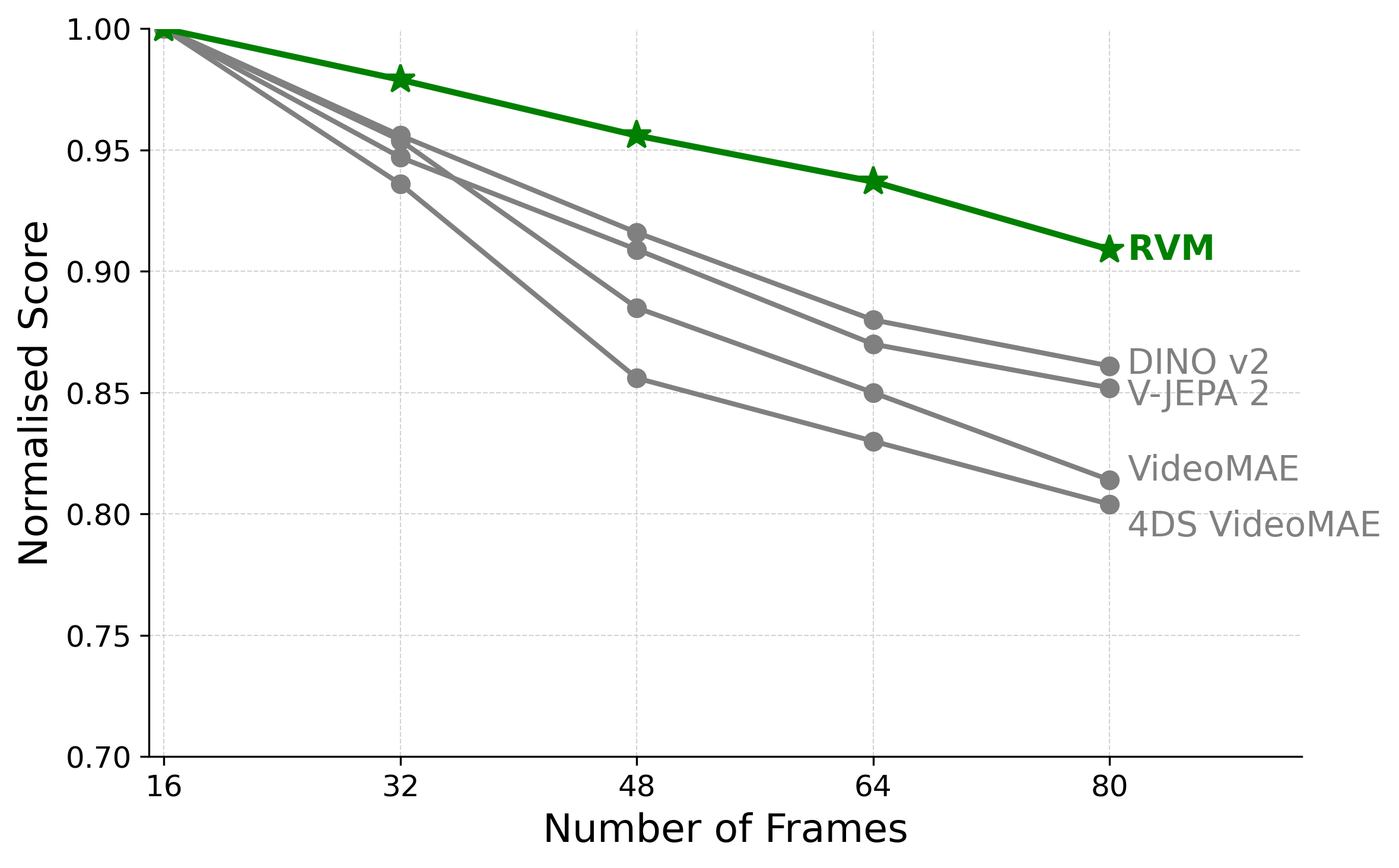}
     \caption{\textbf{RVM features are uniquely stable over long timescales.} We measure temporal stability of visual features by looking at label propagation (feature correspondence) on videos with increasing numbers of frames from the DAVIS 2017 benchmark. RVM performance decays substantially less for long sequences than other SoTA video and image models.}
    \label{fig:stabledavis}
\vspace{-1.5em}
\end{figure}

We compare the stability of features generated by different models over extended time horizons. To do this, we utilize the DAVIS segmentation task, specifically filtering the test dataset to include only videos exceeding 80 frames in length. We then evaluate and compare the tracking performance of the models at intervals of 16, 32, 48, 64, and 80 frames.

Figure \ref{fig:stabledavis} illustrates label-propagation performance as a function of frame count. Results are normalized to each model's performance on 16 frames. As expected, all models perform worse as the time horizon increases. However, \model demonstrates a significantly slower decline in performance, outperforming all other video models as well as strong image-based baselines like DINOv2. This indicates that the recurrent core successfully retains temporally useful information to support long-range correspondence. This result is particularly notable given that \model is trained with only a 4-frame horizon. In contrast, video models that process video in independent blocks, such as VideoMAE, degrade much faster as the number of frames increases, highlighting the critical importance of carrying state across long intervals. As an added benefit, we also note that as the number of frames processed grows, RVM demonstrates significant latency benefits compared with chunked video models. As seen in the Supplementary Material, the recurrent temporal aggregation shows linear latency scaling as opposed to the quadratic scaling of models that use full spatio-temporal self-attention.

\subsection{Ablations}

To better understand the contributions of different model components, we conduct an extensive ablation analysis. All ablation experiments are performed using the Small (S) version of the model, trained on 500M examples (see Table \ref{tab:ablations_1}). We specifically investigate the importance of the time aggregation architecture, the number of source frames, and the scaling behavior with respect to training data size. Full details of ablations can be found in the Supplementary Material.
\vspace{-1.5em}

\paragraph{Number of source frames}
We train the model with 1, 2, and 4 source frames while keeping all other settings constant. The single-frame case serves as an "apples-to-apples" comparison with SiamMAE, controlling for the additional RNN layers and other training differences. With two source frames, the model can capture constant velocity motion, though higher-order dynamics (like acceleration) remain out of reach. We observe a consistent improvement in performance across all tasks as the number of source frames increases from 1 to 2, and further to 4.
\vspace{-1em}
\paragraph{Encoder architecture}
We compare our proposed RNN temporal aggregator against a classic self-attention Transformer. To ensure a fair comparison, we use a patch size of $1\times 16 \times 16$ and match the number of layers and parameters to the RNN core. It is worth noting that the full self-attention mechanism incurs a significantly higher computational cost (FLOPs) compared to the RNN. As shown in the results, the RNN approach is not only more efficient but also performs favorably compared to the self-attention alternative.
\vspace{-1em}
\paragraph{Number of training examples}
Finally, we evaluate how the model's performance scales with the amount of training data. We train four different models on 250M, 500M, 1B, and 2B data samples, respectively. We scale the learning schedules according to the data volume while keeping all other hyperparameters constant. We observe that despite using a relatively small model (34M parameters), our approach continues to benefit from additional data without exhibiting signs of overfitting.

\begin{table}[t]
\centering
\subfloat[
\textbf{Number of source frames.} Having more source frames helps performance. Note that 1 source frame is an apples-to-apples comparison to SiamMAE \citep{gupta2023siamese} with exactly the same data, code and architecture.
\label{tab:source_frames}
]{
\centering
\begin{minipage}{\columnwidth}
\begin{center}
\scriptsize
\tablestyle{3pt}{1.2}
\begin{tabular}{cccc}
num frames & SSv2($\uparrow$) & Kinetics($\uparrow$) & ScanNet($\downarrow$) \\ 
\shline
1 & 41.020 & 39.280 & 1.596 \\
2 & 47.230 & 39.020 & 1.620 \\
\baseline{\textbf{4}} & \baseline{\textbf{52.34}} & \baseline{\textbf{39.72}} & \baseline{\textbf{1.50}} \\
\end{tabular}
\end{center}
\end{minipage}
}
\hspace{1em}
\subfloat[
\textbf{Encoder architecture.} The RNN performs better than full self-attention, while being more computational efficient (see text).
\label{tab:temporal}
]{
\centering
\begin{minipage}{\columnwidth}
\begin{center}
\scriptsize
\tablestyle{3pt}{1.2}
\begin{tabular}{cccc}
aggregator & SSv2($\uparrow$) & Kinetics($\uparrow$) & ScanNet($\downarrow$) \\ 
\shline
SA & 49.1 & 39.6 & 1.505 \\
\baseline{\textbf{RNN}} & \baseline{\textbf{52.34}} & \baseline{\textbf{39.72}} & \baseline{\textbf{1.50}} \\
\end{tabular}
\end{center}
\end{minipage}
}
\hspace{1em}
\subfloat[
\textbf{Number of training examples and steps.} Even with a small model, our approach benefits from more data and training.
\label{tab:num_examples}
]{
\centering
\begin{minipage}{\columnwidth}
\begin{center}
\scriptsize
\tablestyle{3pt}{1.2}
\begin{tabular}{cccc}
num steps & SSv2($\uparrow$) & Kinetics($\uparrow$) & ScanNet($\downarrow$) \\ 
\shline

250M & 46.38 & 33.84 & 1.75 \\
\baseline{{500M}} & \baseline{{52.34}} & \baseline{{39.72}} & \baseline{{1.50}} \\
1B & 55.09 &	44.33 &	1.32 \\
2B & \textbf{57.20} &	\textbf{47.70} &	\textbf{1.200} \\

\end{tabular}
\end{center}
\end{minipage}
}

\vspace{-.1em}
\caption{\textbf{\model ablation experiments}. We ablate some of the components of the model. We show that (a) using more source frames significantly improves results. (b) Using an RNN to aggregate information across time instead of full self-attention is beneficial, especially with tasks that require motion understanding like SSv2 and (c) that the model benefits from more data and training. Default settings for the ablation are marked in \colorbox{baselinecolor}{blue}.}
\label{tab:ablations_1} 
\vspace{-.2in}
\end{table}

\subsection{Qualitative Evaluation}
We begin by qualitatively evaluating the features learned by our model. Using a trained Large (L) model, we unroll it over various test sequences and aggregate the features across time. First, we observe that although the model was trained on only 4 source frames, it generalizes well to much longer sequences without stability issues.

\begin{figure}
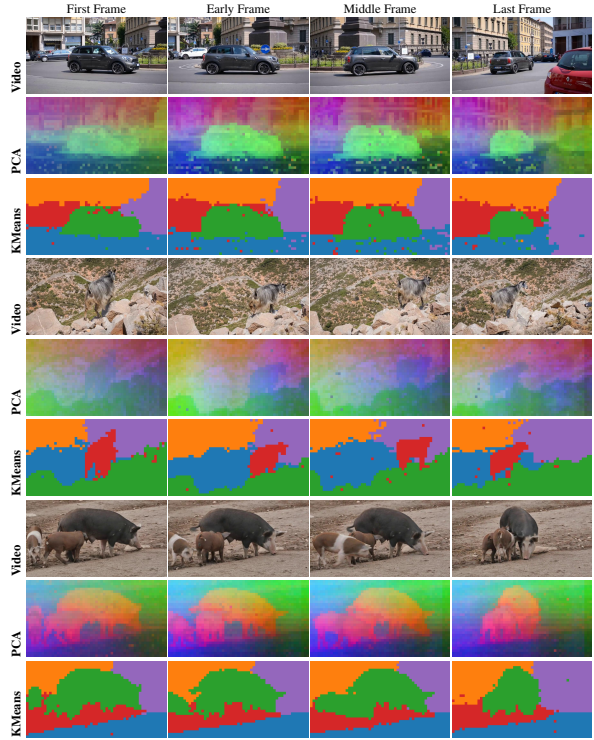

\centering
\newcommand{\imagerow}[7]{%
  \rotatebox{90}{\textbf{#1}} &
  \includegraphics[width=0.22\textwidth]{figures/davis/#2/#3_#4.pdf} &
  \includegraphics[width=0.22\textwidth]{figures/davis/#2/#3_#5.pdf} &
  \includegraphics[width=0.22\textwidth]{figures/davis/#2/#3_#6.pdf} &
  \includegraphics[width=0.22\textwidth]{figures/davis/#2/#3_#7.pdf} \\
}
\resizebox{0.95\linewidth}{!}{%
\begin{tabular}{l @{\hspace{0.5em}} c @{\hspace{0.2em}} c @{\hspace{0.2em}} c @{\hspace{0.2em}} c}
  & First Frame & Early Frame & Middle Frame & Last Frame \\
  
  \imagerow{Video}{frames}{car-roundabout}{0}{18}{37}{74}  
  \imagerow{PCA}{rvm}{car-roundabout_pca}{0}{18}{37}{74}
  \imagerow{KMeans}{rvm}{car-roundabout_kmeans5}{0}{18}{37}{74}
  \imagerow{Video}{frames}{goat}{0}{22}{45}{89}  
  \imagerow{PCA}{rvm}{goat_pca}{0}{22}{45}{89}
  \imagerow{KMeans}{rvm}{goat_kmeans5}{0}{22}{45}{89}
 
  \imagerow{Video}{frames}{pigs}{0}{19}{39}{78}  
  \imagerow{PCA}{rvm}{pigs_pca}{0}{19}{39}{78}
  \imagerow{KMeans}{rvm}{pigs_kmeans5}{0}{19}{39}{78}
\end{tabular}%
}
\caption{\textbf{PCA and K-means of \model features unrolled on unseen videos.} Despite being trained on only 4 frames the model generalizes to long sequences and unrolls stably over long time horizons. As can be seen, the model learns to extract meaningful features from videos.}
\label{fig:rvm_pca_kmeans}
\vspace{-0.8em}
\end{figure}

\begin{figure}
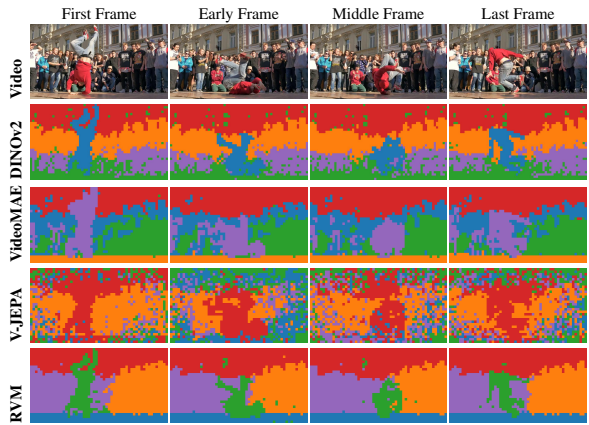

\centering
\newcommand{\imagerow}[7]{%
  \rotatebox{90}{\textbf{#1}} &
  \includegraphics[width=0.17\textwidth]{figures/davis/#2/#3_#4.pdf} &
  \includegraphics[width=0.17\textwidth]{figures/davis/#2/#3_#5.pdf} &
  \includegraphics[width=0.17\textwidth]{figures/davis/#2/#3_#6.pdf} &
  \includegraphics[width=0.17\textwidth]{figures/davis/#2/#3_#7.pdf} \\
}
\resizebox{0.95\linewidth}{!}{%
\begin{tabular}{l @{\hspace{0.5em}} c @{\hspace{0.2em}} c @{\hspace{0.2em}} c @{\hspace{0.2em}} c}
  & First Frame & Early Frame & Middle Frame & Last Frame \\
  \imagerow{Video}{frames}{breakdance}{0}{21}{42}{83}

  \imagerow{DINOv2}{dino2}{breakdance_kmeans5}{0}{21}{42}{83}
  \imagerow{VideoMAE}{videomae}{breakdance_kmeans5}{0}{21}{42}{83}
  
  \imagerow{V-JEPA}{vjepa}{breakdance_kmeans5}{0}{21}{42}{83}

  \imagerow{RVM}{rvm}{breakdance_kmeans5}{0}{21}{42}{83}
\end{tabular}%
}
\caption{\textbf{KMeans visualization on DAVIS video for various ViT-L/16 models.} Unlike 
\model, other models produce noisy feature maps lacking structure and consistency.}
\label{fig:breakdance_kmeans}
\vspace{-1.5em}
\end{figure}

For our first set of test sequences (Figure \ref{fig:rvm_pca_kmeans}), we visualize the features using Principal Component Analysis (PCA) and K-means clustering. For PCA, we concatenate features from all frames and spatial locations, compute the principal components, and map the top three components to the RGB channels of an image. The results show that the model captures meaningful video structures. Similarly, for K-means (with $K=5$), we cluster the concatenated tokens and visualize the resulting segmentation maps by color-coding each cluster. This demonstrates that the model learns to cluster semantically consistent regions in a self-supervised manner.
Figure~\ref{fig:breakdance_kmeans} show K-means clustering for other models. \model produces comparatively clean and stable features.

To show that the model learns meaningful \textit{motion} representation, we test the model on a classic stimulus of a solid white noise square moving on top a static white noise background. This stimulus is interesting because each frame independently is just a white noise image and contains no meaningful structure (as can be seen in Figure~\ref{fig:noise_video_kmeans} top, compare to the video version in the supplementary material). Hence, \textit{image} encoders like DINO or SiamMAE can not extract any useful information from these (Figure~\ref{fig:noise_video_kmeans} second row). \model however is able to "see" the resulting structure. Other video models can also capture the underlying structure, but due to their limited temporal support window, they are unable to provide stable features across the whole sequence (note the cluster reassignments in Figure~\ref{fig:noise_video_kmeans} for VideoMAE).

\begin{figure}
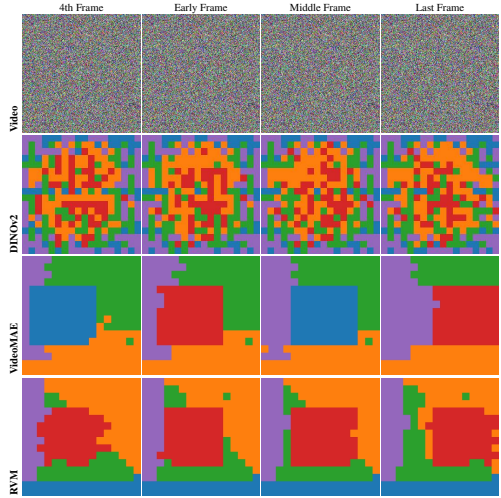

\centering
\newcommand{\imagerow}[7]{%
  \rotatebox{90}{\textbf{#1}} &
  \includegraphics[width=0.22\textwidth]{figures/noise/#2/#3_#4.pdf} &
  \includegraphics[width=0.22\textwidth]{figures/noise/#2/#3_#5.pdf} &
  \includegraphics[width=0.22\textwidth]{figures/noise/#2/#3_#6.pdf} &
  \includegraphics[width=0.22\textwidth]{figures/noise/#2/#3_#7.pdf} \\
}
\resizebox{0.8\linewidth}{!}{%
\begin{tabular}{l @{\hspace{0.45em}} c @{\hspace{0.2em}} c @{\hspace{0.2em}} c @{\hspace{0.2em}} c}
  & 4th Frame & Early Frame & Middle Frame & Last Frame \\
  \imagerow{Video}{frames}{noise_video}{0}{7}{14}{27}
  \imagerow{DINOv2}{dino2}{noise_kmeans5}{0}{7}{14}{27}
  \imagerow{VideoMAE}{videomae}{noise_kmeans5}{0}{7}{14}{27}
  \imagerow{RVM}{rvm}{noise_kmeans5}{0}{7}{14}{27}
\end{tabular}%
}
\caption{\textbf{Detecting a white noise square moving on a white noise background.} From top to bottom: input sequence, \model K-means visualization, an example feature map. Note that each frame independently in the input sequence is a white noise image, and thus image models like DINO or Siam-MAE cannot extract any useful information from these. \model however can integrate temporal information and "see" the moving square. It is highly recommended to watch the video which can be found in the supplementary material. All models use the same ViT-L-16 backbone.\vspace{-1em}}
\label{fig:noise_video_kmeans}
\end{figure}

\section{Limitations}

While RVM sets a new frontier for parameter efficiency and enables linear scaling for long-context inference, this recurrent design incurs specific trade-offs. First, unlike spatio-temporal models such as VideoMAE that patchify across time to reduce token counts, RVM processes frames sequentially. This makes RVM computationally heavier for very short sequences where the benefits of recurrence are less pronounced. Second, training requires back-propagation through time with a ViT encoder at every step, which is memory-intensive. Finally, both a benefit and limitation is that we have yet to find the data saturation point for these models. In this work, we train with 2B clips but find that performance continues to improve with more data. It would be beneficial to establish more formal scaling laws for \model so that we can more efficiently allocate compute.

\section{Conclusion}
We present Recurrent Video Masked-Autoencoders (RVM), a novel framework that leverages recurrent computation as a way to integrate temporal information in self-supervised video representation learning. By coupling an asymmetric masking autonecoder style training objective with a transformer-based recurrent core, \model effectively aggregates information over time to learn "generalist" visual representations. Our results demonstrate that \model presents a unique advance in the landscape of current vision models: it matches or exceeds the spatio-temporal capabilities of video-centric models (e.g., VideoMAE, V-JEPA) while retaining the dense, spatial and geometric understanding properites of strong frame-centric models (e.g., DINOv2). Furthermore, \model introduces a way to train strong small models, trained without the need for knowledge-distillation, that exhibit up to $30\times$ greater parameter efficiency for the same averaged performance. Finally, we find that this recurrent architecture exhibits superior feature stability over long temporal horizons compared to state-of-the-art ``video model'' baselines. In sum, our work suggests that bringing back recurrent video processing with a simple pixel-level training objective may be sufficient for learning strong visual models from natural video data without the need for extra tricks like strong augmentation, EMA networks, regularizers etc. Future work will explore further scaling our method and evaluating it in the context of multi-modal and world modeling tasks like robotic control.

\section*{Acknowledgment}
We thank Goker Erdogan, Viorica P{\u{a}}tr{\u{a}}ucean, Aravindh Mahendran, Miki Rubinstein, Dilara Gokay, Junlin Zhang and Joseph Heyward for helpful discussions and support.

{
    \small
    \bibliographystyle{ieeenat_fullname}
    \bibliography{main}
}
\clearpage
\clearpage
\maketitlesupplementary
\section{Training data details}
\label{supp:training}
We use a data mixture very similar to the one proposed in \cite{assran2025v}, consisting of only data from publically available video datasets. However, we do not apply any extra curation to these datasets and critically don't rely on ImageNet for additional image-level data as so many prior works do: 

\begin{table}[ht]
\centering
\resizebox{\columnwidth}{!}{\begin{tabular}{lccccc}
\toprule
\multicolumn{1}{c}{Source} & Samples & Type & FPS & Apply Curation & Weight \\ 
\midrule
SSv2 \citep{goyal2017something} & 168K & EgoVideo & 25 & No & 0.056 \\
Kinetics-700 \citep{carreira2019short} & 733K & ExoVideo & 25 & No & 0.188 \\
Howto100M \citep{miech2019howto100m} & 1.1M & ExoVideo & 10 & No & 0.318 \\
YT8M \citep{abu2016youtube} & 3.3M & ExoVideo & 10 & No & 0.188 \\
YT-BoundingBoxes \citep{real2017youtube} & 380K & ExoVideo & 10 & No & 0.250 \\ 
\bottomrule
\end{tabular}}
\caption{Dataset usage and statistics. We use only video datasets. While we apply no curation ourselves to any of these datasets, the original dataset construction for many of these datasets (except YT8M) did involve significant curation. For YT8M we utilize available clips- a significant number of clips from the original dataset can no longer be accessed.}
\label{tab:datasets}
\end{table}

\noindent A training batch consists of selecting videos with the mixture weights specified in Table \ref{tab:datasets}. Each clip from a given datasets is a 64 frame clip (which corresponds to different durations because of the differing fps for each source). We use the first 4 consecutive frames for the source frames and sample a target frame with a uniform temporal gap of 4 to 48 frames. For each video clip we apply the following augmentations: 
\begin{enumerate}
\item Video-level RandomHorizontalFlipping ($p=0.5$)
\item Frame-level RandomResizedCrop with $\text{scale}=(0.3, 1.0)$ and $\text{aspect ratio} = (0.75, 1.25)$, using bicubic interpolation. 
\end{enumerate}

\section{Architecture details}
\label{supp:architecture}
We provide the network architecture details for each model component in Tables \ref{tab:arch_details}
 and \ref{tab:decoder_details}.
\begin{table}[ht]
    \centering
    \caption{RVM Architecture Variants. We scale the Encoder and RNN core across four sizes (S, B, L, H). The Encoder follows standard ViT specifications \cite{dosovitskiy2020image}. The RNN core dimension matches the encoder embedding dimension.}
    \label{tab:arch_details}
    \resizebox{\linewidth}{!}{
    \begin{tabular}{l|cccc|cc|r}
        \toprule
        \multicolumn{1}{c|}{\textbf{Model}} & \multicolumn{4}{c|}{\textbf{ViT Encoder}} & \multicolumn{2}{c|}{\textbf{RNN Core}} & \multicolumn{1}{c}{\textbf{Total}} \\
         & \textbf{Embed Dim} & \textbf{Heads} & \textbf{Layers} & \textbf{MLP Ratio} & \textbf{Layers} & \textbf{Heads} & \textbf{Params} \\
        \midrule
        RVM-S & 384 & 6 & 12 & 4.0 & 4 & 8 & 34M \\
        RVM-B & 768 & 12 & 12 & 4.0 & 4 & 12 & 117M \\
        RVM-L & 1024 & 16 & 24 & 4.0 & 4 & 16 & 375M \\
        RVM-H & 1280 & 16 & 32 & 4.0 & 4 & 16 & 743M \\
        \bottomrule
    \end{tabular}
    }
\end{table}
As specified in the main text, \model-S,B,L models are trained for 1M steps (approx. 2B samples). However, we find that larger models do benefit from even longer schedules and thus train our \model-H for 4B steps. 
\begin{table}[ht]
    \centering
    \caption{Decoder Architecture. The decoder is fixed across all model sizes. Each block consists of Cross-Attention (Target-Source), MLP, and Self-Attention layers. Refer to pseudocode in Section \ref{sec:code}}
    \label{tab:decoder_details}
     \resizebox{\linewidth}{!}{
    \begin{tabular}{lc}
        \toprule
        \textbf{Hyperparameter} & \textbf{Value} \\
        \midrule
        Embedding Dimension ($D_{dec}$) & 512 \\
        Number of Heads & 16 \\
        Number of Blocks & 8 \\
        MLP Ratio & 4.0 \\
        \midrule
        \textit{Block Structure} & \\
        1. Cross-Attention & Target (Q) $\leftrightarrow$ Source (K,V) \\
        2. Feed-Forward & MLP \\
        3. Self-Attention & Target (Q,K,V) \\
        \bottomrule
    \end{tabular}
    }
\end{table}

\section{Self attention ablation details}
To ensure a fair comparison between our recurrent temporal aggregation and a full self-attention approach, we minimized differences in the experimental setup. We maintained the exact same encoder, architecture, and hyperparameters. The primary distinction lies in how tokens are prepared for the ViT. In the full self-attention baseline, we patchify and project frames independently but concatenate all resulting tokens along the token axis before feeding them into the ViT (accounting for the extra layers due to the RNN core.). We also augmented the positional embeddings with a time dimension to provide temporal context. This setup is essentially equivalent to using a $1 \times 16 \times 16$ patch size in spatiotemporal models like VideoMAE. All other components, including masking patterns, training objectives, and learning schedules, remained identical to the RNN configuration.
\label{supp:ablataions}

\section{Pseduocode}
\label{sec:code}
\begin{lstlisting}[caption={RVM Recurrent Core Pseudo-code}, label={lst:rvm_code}]
class RVMCell(nn.Module):
    def __init__(self, dim, transformer_block):
        super().__init__()
        self.Tx = transformer_block
        # Update (u) and Reset (r) gate projections
        self.We_u, self.Ws_u = nn.Linear(dim, dim), nn.Linear(dim, dim)
        self.We_r, self.Ws_r = nn.Linear(dim, dim), nn.Linear(dim, dim)

    def forward(self, x_seq):
        """
        x_seq: Sequence of source frame tokens [e_1, ..., e_K]
        Returns: Sequence of refined features [o_1, ..., o_K]
        """
        # Initialize state s_0 to zero
        s = torch.zeros_like(x_seq[0]) 
        outputs = []

        for x in x_seq:
            # 1. Compute Gates (Eq. 1)
            u = torch.sigmoid(self.We_u(x) + self.Ws_u(s))
            r = torch.sigmoid(self.We_r(x) + self.Ws_r(s))

            # 2. Transformer Integration (Eq. 2)
            # Query is current input (x); KV is reset-gated state
            h = self.Tx(query=x, kv=r * s)

            # 3. State Update (Eq. 2)
            s = (1 - u) * s + u * h
            
            # 4. Output (Eq. 3)
            outputs.append(s)

        return torch.stack(outputs)
\end{lstlisting}

\begin{lstlisting}[caption={Cross Attention block used in RNN core and decoder}, label={lst:xa_code}]
class TransformerBlock(nn.Module):
    def __init__(self, dim, num_heads):
        super().__init__()
        self.ln1 = nn.LayerNorm(dim)
        self.self_attn = nn.MultiheadAttention(dim, num_heads)
        
        self.ln2 = nn.LayerNorm(dim)
        self.cross_attn = nn.MultiheadAttention(dim, num_heads)
        
        self.ln3 = nn.LayerNorm(dim)
        self.mlp = MLP(dim) # Standard Feed Forward

    def forward(self, x, mem):
        """
        x: Current frame tokens (Query) [cite: 350]
        mem: Gated previous state (Key/Value) [cite: 350, 476]
        """
        # 1. Self-Attention (Intra-frame mixing)
        # Using pre-normalization [cite: 341]
        x = x + self.self_attn(query=self.ln1(x), 
                               key=self.ln1(x), 
                               value=self.ln1(x))[0]

        # 2. Feed Forward
        x = x + self.mlp(self.ln3(x))

        # 3. Cross-Attention (Temporal integration)
        # Queries from current frame, Keys/Values from history
        x = x + self.cross_attn(query=self.ln2(x), 
                                key=mem, 
                                value=mem)[0]
        
        return x
\end{lstlisting}

\section{Evaluation Details}
\label{supp:eval}

To comprehensively assess the capabilities of Recurrent Video Masked Autoencoders (RVM), we evaluate the model across a broad spectrum of 8 diverse datasets covering high-level semantics, low-level geometry, and temporal correspondence. Our evaluation suite encompasses distinct visual tasks including action recognition (SSv2~\citep{ssv2}, Kinetics-700~\citep{kinetics}), monocular depth estimation (ScanNet~\citep{dai2017scannet}), and fine-grained motion tracking (Perception Test~\citep{patraucean2023perception}, Waymo Open~\citep{sun2020waymo}). Additionally, we probe the spatio-temporal consistency of the learned features through non-parametric nearest-neighbor label propagation on the DAVIS-2017~\citep{pont20172017}, JHMDB~\citep{jhuang2013towards}, and VIP~\citep{zhou2018adaptive} benchmarks. This exhaustive protocol ensures a holistic comparison against existing state-of-the-art video and image foundation models.

\subsection{Downstream tasks}
We adopt the rigorous evaluation protocol of \citet{carreira2024scaling}, attaching lightweight attention-based readouts to frozen backbones.

\begin{itemize}
    \item \textbf{SSv2 action recognition} \citep{ssv2}: A fine-grained dataset requiring temporal understanding. We process 16-frame clips at $224 \times 224$ resolution with a stride of 2. The readout employs a cross-attention layer with 768 channels and 12 heads, using a single learned query to pool representations before the final linear classifier. Training involves color augmentation (brightness, contrast, saturation, hue) and random grayscale conversion. We report top-1 accuracy (\%).

    \item \textbf{Kinetics-700-2020 action recognition} \citep{kinetics}: A large-scale benchmark for broad action understanding. Similar to SSv2, we use 16-frame clips with a stride of 2. The readout is larger, utilizing 1024 channels and 16 heads with a single learned query. For evaluation, we average predictions over 7 linearly spaced temporal clips per video. We report top-1 accuracy (\%).

    \item \textbf{ScanNet depth estimation} \citep{dai2017scannet}: Evaluates geometric understanding on indoor RGB-D videos. We input 16 RGB frames and predict dense depth maps. The readout uses cross-attention (1024 channels, 16 heads) where queries are learned features corresponding to each $2\times8\times8$ patch. The model minimizes an $L_2$ loss on log-scale depth. Performance is measured by Absolute Relative Error (AbsRel).

    \item \textbf{Perception Test point tracking} \citep{patraucean2023perception}: Measures fine-grained long-term motion tracking. The readout uses cross-attention (1024 channels, 8 heads) where queries are derived from the initial point positions embedded via Fourier features. The model predicts position, visibility, and uncertainty for each track. Following \citet{carreira2024scaling}, the readout is trained on the synthetic Kubric MOVI-E dataset \citep{greff2022kubric} before evaluating on the real-world Perception Test. We report Average Jaccard (AJ).

    \item \textbf{Waymo Open object tracking} \citep{sun2020waymo}: Assesses object-level motion consistency in driving scenarios. We track 2D bounding boxes over 16-frame clips ($256 \times 256$ resolution). The readout employs cross-attention (1024 channels, 4 heads) with queries formed from the initial bounding box coordinates. We report mean Intersection-over-Union (mIoU).
\end{itemize}

\begin{table}[t]
    \centering
    \caption{\textbf{Downstream Task Readout Hyperparameters.} Summary of the attention-based readout configurations used for each task, following the protocol of \citet{carreira2024scaling}. All readouts use a Cross-Attention (CA) mechanism on top of the frozen backbone features.}
    \label{tab:readout_params}
    \resizebox{\linewidth}{!}{
    \begin{tabular}{lcccc}
        \toprule
        \textbf{Task} & \textbf{Input Shape} & \textbf{Readout Arch} & \textbf{Heads / Channels} & \textbf{Query Type} \\
        \midrule
        SSv2 & $16 \times 224^2$ & Cross-Attn & $12$ / $768$ & Learned Vector \\
        Kinetics-700 & $16 \times 224^2$ & Cross-Attn & $16$ / $1024$ & Learned Vector \\
        ScanNet & $16 \times 224^2$ & Cross-Attn & $16$ / $1024$ & Learned Vector \\
        Perception Test & $16 \times 224^2$ & Cross-Attn & $8$ / $1024$ & Point Coord. + Fourier \\
        Waymo Open & $16 \times 256^2$ & Cross-Attn & $4$ / $1024$ & Box Coord. + Fourier \\
        \bottomrule
    \end{tabular}
    }
\end{table}

\subsection{Nearest-neighbor tasks}  
Unlike read-out classification, this protocol directly probes whether the pre-trained features encode spatially and temporally consistent information without any task-specific training.
\begin{itemize}
    \item \textbf{DAVIS-2017 video segmentation tracking} \citep{pont20172017}: A video object segmentation benchmark with diverse object categories and complex motion. The task is to propagate ground-truth instance masks provided in the first frame across subsequent frames. We adopt the non-parametric label propagation algorithm of \citet{jabri2020space} that considers the similarity between patch features across frames, using 480p resolution with patch sizes 14/16 matched across models. Like DINO \citep{caron2021emerging}, performance is reported in the standard $\mathcal{J}$\&$\mathcal{F}$-mean metric, which combines region similarity ($\mathcal{J}$) and contour accuracy ($\mathcal{F}$) \citep{perazzi2016benchmark}, computed at the native resolution of the videos. 
    \item \textbf{JHMDB human keypoint tracking} \citep{jhuang2013towards}: A dataset of short video clips for human pose estimation and action understanding. We follow the setup of \citet{li2019joint}, using 320$\times$320 video resolution and a single context frame, and report PCK@0.1.
    \item \textbf{VIP human part tracking} \citep{xu2018video}: A video instance segmentation benchmark requiring pixel-level separation of multiple moving instances. \citep{zhou2018adaptive} requires dense propagation of semantic part masks across long human-centric videos, with up to 20 different human part categories and durations of 120 seconds. Following the protocol of \citet{li2019joint}, we evaluate at 448$\times$880 resolution using a single context frame.
\end{itemize}

\begin{table}[t]
    \centering
    \caption{\textbf{Label Propagation Evaluation Protocols.} Summary of hyperparameters used across DAVIS, JHMDB, and VIP tasks. The models share the same temperature and memory bank size, differing mainly in resolution and $k$-NN retrieval count.}
    \label{tab:eval_protocols}
    \resizebox{\linewidth}{!}{
    \begin{tabular}{lccc}
        \toprule
        \textbf{Parameter} & \textbf{DAVIS-2017} & \textbf{JHMDB} & \textbf{VIP} \\
        \midrule
        Task & VOS & Keypoint Tracking & Part Propagation \\
        Resolution & $480 \times 880$ & $320 \times 320$ & $448 \times 880$ \\
        Metric & $\mathcal{J}\&\mathcal{F}$ Mean & PCK@0.1 & mIoU \\
        Top-$k$ ($k$) & 7 & 7 & 10 \\
        \midrule
        Algorithm & \multicolumn{3}{c}{Non-parametric Label Propagation \cite{jabri2020space}} \\
        Temperature ($\tau$) & \multicolumn{3}{c}{0.7} \\
        Context Frames & \multicolumn{3}{c}{20} \\
        Search Radius & \multicolumn{3}{c}{20} \\
        \bottomrule
    \end{tabular}
    }
\end{table}

\section{Baseline Models}
\textbf{Backbone architectures} We evaluate models including SiamMAE, DINOv2, VideoMAE, VideoMAEv2, V-JEPA, and 4DS. Most baselines use Vision Transformers (ViTs) with spatio-temporal patch tokenization of size $(2,16,16)$, where each token covers two consecutive frames and a $16\times16$ spatial region. Self-attention is applied across all tokens, making computation quadratic in the number of patches. We evaluate models across a wide range of capacities, from ViT-S ($\sim$30M parameters) up to ViT-H ($\sim$700M parameters). The exact configurations, pre-training checkpoints, and architectural details for each model are provided in Table~\ref{tab:model_configs}. Below, we provide a concrete description of each model included in our experiments.

\subsection{VideoMAE and VideoMAEv2}
As a representative video-masked autoencoder, VideoMAE~\cite{he2022masked, feichtenhofer2021large} operates on a standard Vision Transformer (ViT) backbone processing tubelets of size $2\times16\times16$. It employs a high masking ratio and reconstructs normalized pixels of masked regions using a vanilla ViT decoder. Building on this, VideoMAE v2~\cite{wang2023videomae} incorporates a dual masking strategy—masking tokens in both the encoder and decoder—to enhance computational efficiency and scalability. We examine variants ranging from ViT-B (30M parameters) to the billion-parameter ViT-g. These models are typically pretrained on Kinetics-400, with v2 leveraging a progressive training schedule on a massive mixed dataset of public videos. We utilize the official checkpoints respectively\footnote{https://github.com/MCG-NJU/VideoMAE}\textsuperscript{,}\footnote{https://github.com/OpenGVLab/VideoMAEv2}.

\subsection{V-JEPA} 
The V-JEPA family utilizes a Joint-Embedding Predictive Architecture (JEPA) to learn semantic video representations. A ViT encoder~\cite{dosovitskiy2020image} processes 16-frame inputs (resolution $224 \times 224$) decomposed into $2\times16\times16$ patches. Unlike generative approaches, V-JEPA is trained to predict the latent representation of a target video signal $y$ from a context $x$ (a heavily masked version of $y$) by minimizing the $L_1$ distance in feature space. The target encoder is updated via an exponential moving average (EMA) of the context encoder. We evaluate ViT-L ($\sim$300M) and ViT-H ($\sim$600M) variants pretrained for 90k iterations on VideoMix2M, a compilation of HowTo100M, Kinetics-400/600/700 (K710)~\cite{kay2017kineticshumanactionvideo}, and Something-Something-v2~\cite{goyal2017something}. We use the official model checkpoints\footnote{https://github.com/facebookresearch/jepa}.

\subsection{DINOv2}
DINOv2~\cite{oquab2023dinov2} adapts the DINO self-distillation framework to large-scale data. It processes video frames independently as images using a ViT~\cite{dosovitskiy2020image} with patch size 14, yielding a feature grid of $16\times16\times16$ for a 16-frame input. The training objective combines contrastive and distillation losses at both the image and patch levels, supported by sophisticated data curation and regularization techniques. We utilize the official pre-trained checkpoints \footnote{https://github.com/facebookresearch/dinov2} for ViT-L (307M) and ViT-g (1.1B), which were trained for 625k steps with a batch size of 3,072, applying the model frame-by-frame to generate video features.

\subsection{4DS}
The 4DS framework \cite{carreira2024scaling} simplifies the masked autoencoding paradigm (SimpleMAE) by discarding the separate lightweight decoder in favor of using the last few self-attention blocks of the encoder for reconstruction. It employs a standard ViT with $2\times16\times16$ tokenization but opts for a random masking strategy (95\% ratio) over tube masking. The model minimizes the $L_2$ reconstruction loss on RGB values across all patches—both masked and unmasked—without target normalization. We evaluate a ViT-B variant pretrained on a massive corpus of 170 million web videos (1 billion clips). We use the official checkpoints \footnote{https://github.com/google-deepmind/representations4d}.

\begin{table}[ht]
    \centering
    \caption{\textbf{Summary of Pre-trained Models used for Nearest Neighbor Evaluation.} We report the architecture, patch size ($P$), embedding dimension ($D$), depth ($L$), number of heads ($H$), and the pre-training dataset for each checkpoint used.}
    \label{tab:model_configs}
    \resizebox{\linewidth}{!}{%
    \begin{tabular}{lccccc}
        \toprule
        \textbf{Model} & \textbf{Arch.} & \textbf{$P$} & \textbf{$D$} & \textbf{$L$} & \textbf{$H$}  \\
        \midrule
        VideoMAE & ViT-L & $16{\times}16$ & 1024 & 24 & 16 \\
        VideoMAE v2 & ViT-L & $16{\times}16$ & 1024 & 24 & 16 \\
        V-JEPA & ViT-L & $16{\times}16$ & 1024 & 24 & 16 \\
       
        DINOv2 & ViT-L & $14{\times}14$ & 1024 & 24 & 16 \\
        4DS-VideoMAE & ViT-B & $16{\times}16$ & 768 & 12 & 12 \\
        RVM & ViT-L & $16{\times}16$ & 1024 & 24 & 16  \\        
        
        \bottomrule
    \end{tabular}
    }
\end{table}

\section{Video model latency vs. number of frames processed}
\begin{figure}
\centering
\includegraphics[width=0.9\columnwidth]{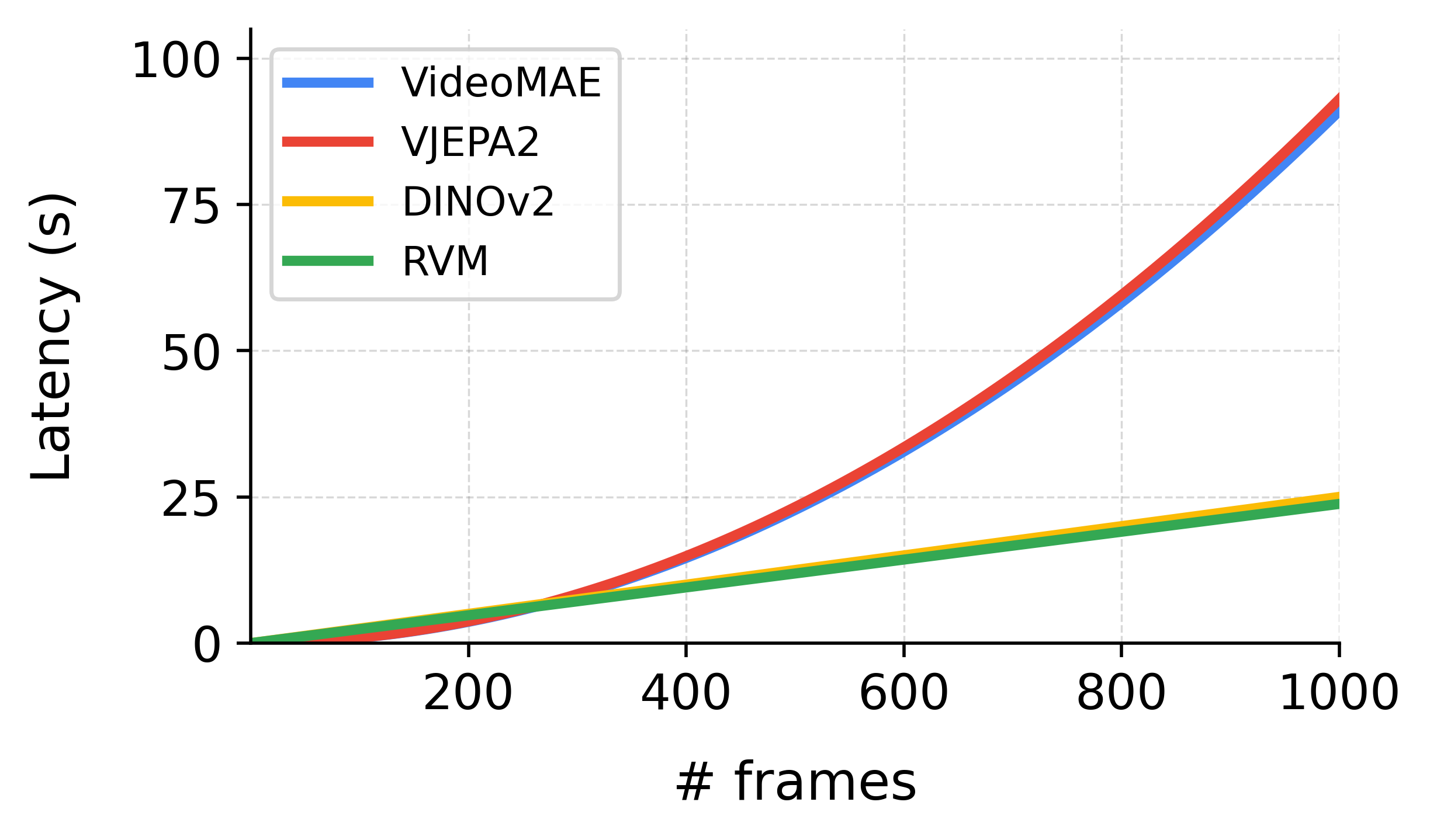}
\caption{Traditional video encoders suffer from quadratic latency growth with number of frames due to spatio-temporal self-attention. RVM still achieves strong temporal integration properties while having latency gros linearly with the number of frames, like pure frame-based models (e.g. DinoV2).}
\label{fig:supp_latency}
\end{figure}

We see in Fig. \ref{fig:supp_latency} that RVM exhibits linear inference latency scaling with number of frames. This is akin to frame-based models and is more amenable to streaming than chunked video models that grow latency quadratically w.r.t to the numebr of frames.

\section{Additional results for \model-B scale}
We include in Table \ref{supp:tab-rvm} additional results for the B model scale. While the B model scale doesn't outperform the S model by a large margin, it is still better than comparable image (SiamMAE) and video (VideoMAE-B/4DS-B) baselines. 
\begin{table*}
\resizebox{\textwidth}{!}{
\begin{tabular}{lcccccccccc}
\toprule
& & \multicolumn{4}{c}{Spatio-temporal Tasks} & \multicolumn{4}{c}{Spatial Tasks} & \\
\cmidrule(r){3-6} \cmidrule(r){7-10}
Model & Size(M) & SSv2 & Kinetics & Waymo & PT & ScanNet & DAVIS & JHMDB & VIP \\
& & Acc. ($\uparrow$\%) & Acc. ($\uparrow$\%) & mIoU ($\uparrow$) & AJ ($\uparrow$) & AbsRel ($\downarrow$) & J\&F ($\uparrow$\%) & PCK@0.1 ($\uparrow$\%) & mIoU ($\uparrow$)   \\
\midrule
SiamMAE-B & 85 & -- & -- & -- & -- & -- & 62.8 & 47.2 & \textbf{38.4}  \\
VideoMAE-B & 87 & 52.3 & 38.9 & \textbf{73.1} & \textbf{79.2} & 1.50 & 50.6 & 44.7 & 19.6  \\
4DS-B & 91 & 49.6 & 35.7 & 72.7 & 78.9 & 1.65 & -- & -- & --  \\
\baselinerow \model-B & 117 & \textbf{61.4} & \textbf{53.1} & 71.1 & 74.5 & \textbf{1.08} & \textbf{63.9} & \textbf{49.4} & 35.8  \\
\bottomrule
\end{tabular}}
\caption{Additional results for the RVM-B model scale and a few competitive baselines.}
\label{supp:tab-rvm}
\end{table*}

\section{Additional Qualitative Results}
\label{supp:qualitative}

To probe the spatiotemporal structure of the learned representations, we visualize the dense feature maps extracted from the frozen backbone of each model. We employ two standard dimensionality reduction techniques:

\paragraph{Principal Component Analysis (PCA).} We compute the top-3 principal components of the flattened feature tokens across the entire video volume. These components are whitened and mapped to RGB color channels. This visualization highlights the global structure and smoothness of the feature space, revealing whether the model separates foreground motion from the background.

\paragraph{K-Means Clustering.} We apply K-means clustering with $k=5$ clusters (initialized via k-means++) to the feature descriptors. Each cluster is assigned a distinct color to generate a segmentation mask. This acts as a proxy for semantic understanding, testing whether spatially coherent regions (e.g., an object's parts) are grouped together and whether these assignments remain temporally consistent across frames.

\begin{figure}
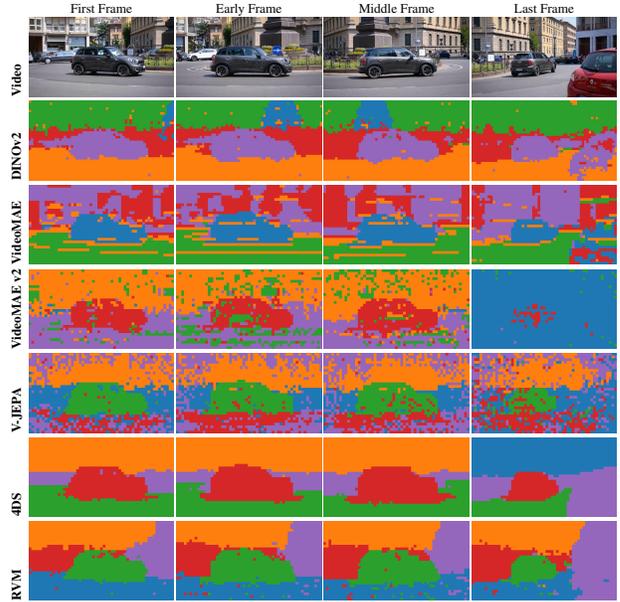

\centering
\newcommand{\imagerow}[7]{%
  \rotatebox{90}{\textbf{#1}} &
  \includegraphics[width=0.22\textwidth]{figures/davis/#2/#3_#4.pdf} &
  \includegraphics[width=0.22\textwidth]{figures/davis/#2/#3_#5.pdf} &
  \includegraphics[width=0.22\textwidth]{figures/davis/#2/#3_#6.pdf} &
  \includegraphics[width=0.22\textwidth]{figures/davis/#2/#3_#7.pdf} \\
}
\resizebox{0.99\linewidth}{!}{%
\begin{tabular}{l @{\hspace{0.5em}} c @{\hspace{0.2em}} c @{\hspace{0.2em}} c @{\hspace{0.2em}} c}
  & First Frame & Early Frame & Middle Frame & Last Frame \\
  \imagerow{Video}{frames}{car-roundabout}{0}{18}{37}{74}
  \imagerow{DINOv2}{dino2}{car-roundabout_kmeans5}{0}{18}{37}{74}
  \imagerow{VideoMAE}{videomae}{car-roundabout_kmeans5}{0}{18}{37}{74}
  \imagerow{VideoMAE v2}{videomae2}{car-roundabout_kmeans5}{0}{18}{37}{74}
  \imagerow{V-JEPA}{vjepa}{car-roundabout_kmeans5}{0}{18}{37}{74}
  \imagerow{4DS}{4ds}{car-roundabout_kmeans5}{0}{18}{37}{74}
  \imagerow{RVM}{rvm}{car-roundabout_kmeans5}{0}{18}{37}{74}
\end{tabular}%
}
\caption{\textbf{Temporal Stability in Feature Space.} Using K-Means clustering ($k=5$) on the \texttt{car-roundabout} sequence, we observe that \textbf{RVM} (Ours) maintains stable cluster assignments for the moving vehicle and the background throughout the clip. In contrast, \textbf{VideoMAE v2} and \textbf{4DS} exhibits significant temporal discontinuity ("flickering"), failing to track the object or background consistently over time.}
\label{fig:car-roundabout_kmeans}
\end{figure}

\begin{figure}
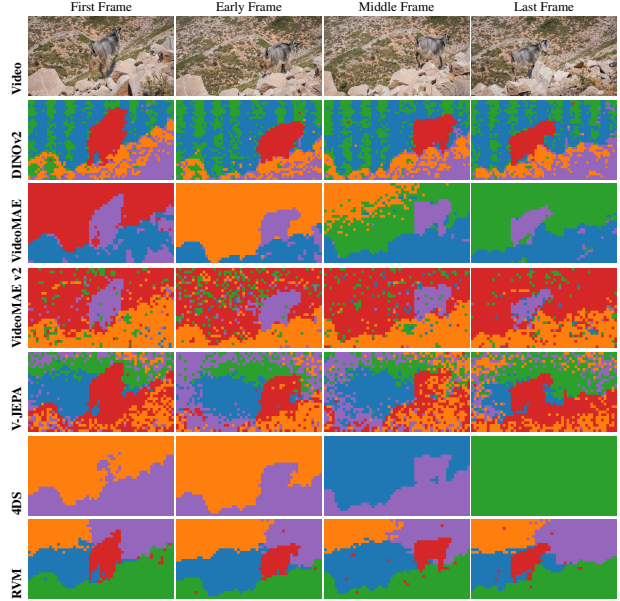

\centering
\newcommand{\imagerow}[7]{%
  \rotatebox{90}{\textbf{#1}} &
  \includegraphics[width=0.22\textwidth]{figures/davis/#2/#3_#4.pdf} &
  \includegraphics[width=0.22\textwidth]{figures/davis/#2/#3_#5.pdf} &
  \includegraphics[width=0.22\textwidth]{figures/davis/#2/#3_#6.pdf} &
  \includegraphics[width=0.22\textwidth]{figures/davis/#2/#3_#7.pdf} \\
}
\resizebox{0.99\linewidth}{!}{%
\begin{tabular}{l @{\hspace{0.5em}} c @{\hspace{0.2em}} c @{\hspace{0.2em}} c @{\hspace{0.2em}} c}
  & First Frame & Early Frame & Middle Frame & Last Frame \\
  \imagerow{Video}{frames}{goat}{0}{22}{45}{89}
  \imagerow{DINOv2}{dino2}{goat_kmeans5}{0}{22}{45}{89}
  \imagerow{VideoMAE}{videomae}{goat_kmeans5}{0}{22}{45}{89}
  \imagerow{VideoMAE v2}{videomae2}{goat_kmeans5}{0}{22}{45}{89}
  \imagerow{V-JEPA}{vjepa}{goat_kmeans5}{0}{22}{45}{89}
  \imagerow{4DS}{4ds}{goat_kmeans5}{0}{22}{45}{89}
  \imagerow{RVM}{rvm}{goat_kmeans5}{0}{22}{45}{89}
\end{tabular}%
}
\caption{\textbf{Robust Foreground-Background Segmentation.} In the \texttt{goat} sequence, \textbf{RVM} effectively disentangles the moving animal from the complex environment. While \textbf{4DS} suffers from background confusion, merging the object with the scene, \textbf{RVM} produces clean, spatially coherent segments that adhere strictly to object boundaries. \textbf{DINOv2} segments the object well but fails significantly on the background.}
\label{fig:goat_kmeans}
\end{figure}

\begin{figure}
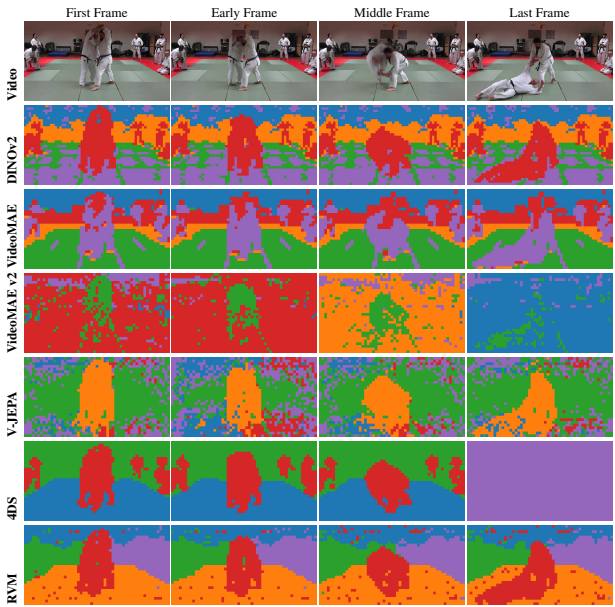

\centering
\newcommand{\imagerow}[7]{%
  \rotatebox{90}{\textbf{#1}} &
  \includegraphics[width=0.22\textwidth]{figures/davis/#2/#3_#4.pdf} &
  \includegraphics[width=0.22\textwidth]{figures/davis/#2/#3_#5.pdf} &
  \includegraphics[width=0.22\textwidth]{figures/davis/#2/#3_#6.pdf} &
  \includegraphics[width=0.22\textwidth]{figures/davis/#2/#3_#7.pdf} \\
}
\resizebox{0.99\linewidth}{!}{%
\begin{tabular}{l @{\hspace{0.5em}} c @{\hspace{0.2em}} c @{\hspace{0.2em}} c @{\hspace{0.2em}} c}
  & First Frame & Early Frame & Middle Frame & Last Frame \\
  \imagerow{Video}{frames}{judo}{0}{8}{17}{33}

  \imagerow{DINOv2}{dino2}{judo_kmeans5}{0}{8}{17}{33}
  \imagerow{VideoMAE}{videomae}{judo_kmeans5}{0}{8}{17}{33}
  \imagerow{VideoMAE v2}{videomae2}{judo_kmeans5}{0}{8}{17}{33}
  \imagerow{V-JEPA}{vjepa}{judo_kmeans5}{0}{8}{17}{33}

  \imagerow{4DS}{4ds}{judo_kmeans5}{0}{8}{17}{33}
  \imagerow{RVM}{rvm}{judo_kmeans5}{0}{8}{17}{33}
\end{tabular}%
}
\caption{\textbf{Motion-Aware Instance Separation.} Visualizing clusters for the \texttt{judo} sequence. \textbf{RVM} preserves the structural integrity of semantic parts while separating moving instances from static ones (foreground vs.\ background human). Notably, it filters out the static background human that \textbf{DINOv2} fails to distinguish.}
\label{fig:judo_kmeans}
\end{figure}

\begin{figure}
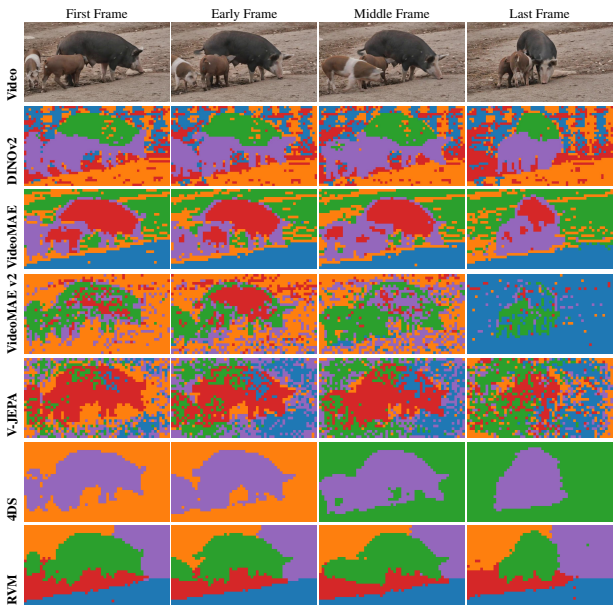

\centering
\newcommand{\imagerow}[7]{%
  \rotatebox{90}{\textbf{#1}} &
  \includegraphics[width=0.22\textwidth]{figures/davis/#2/#3_#4.pdf} &
  \includegraphics[width=0.22\textwidth]{figures/davis/#2/#3_#5.pdf} &
  \includegraphics[width=0.22\textwidth]{figures/davis/#2/#3_#6.pdf} &
  \includegraphics[width=0.22\textwidth]{figures/davis/#2/#3_#7.pdf} \\
}
\resizebox{0.99\linewidth}{!}{%
\begin{tabular}{l @{\hspace{0.5em}} c @{\hspace{0.2em}} c @{\hspace{0.2em}} c @{\hspace{0.2em}} c}
  & First Frame & Early Frame & Middle Frame & Last Frame \\
  \imagerow{Video}{frames}{pigs}{0}{19}{39}{78}

  \imagerow{DINOv2}{dino2}{pigs_kmeans5}{0}{19}{39}{78}
  \imagerow{VideoMAE}{videomae}{pigs_kmeans5}{0}{19}{39}{78}
  \imagerow{VideoMAE v2}{videomae2}{pigs_kmeans5}{0}{19}{39}{78}
  \imagerow{V-JEPA}{vjepa}{pigs_kmeans5}{0}{19}{39}{78}

  \imagerow{4DS}{4ds}{pigs_kmeans5}{0}{19}{39}{78}
  \imagerow{RVM}{rvm}{pigs_kmeans5}{0}{19}{39}{78}
\end{tabular}%
}
\caption{\textbf{Long-Term Consistency under Deformation.} Visual results on the \texttt{pigs} sequence demonstrate \textbf{RVM}'s ability to maintain consistency over time. While \textbf{V-JEPA} exhibits cluster fragmentation, \textbf{RVM} leverages recurrent cues to effectively preserve the identity of semantic parts during non-rigid motion.}
\label{fig:pigs_kmeans}
\end{figure}

\noindent As shown in Figures~\ref{fig:car-roundabout_kmeans}--\ref{fig:pigs_kmeans}, \textbf{RVM} demonstrates remarkable temporal consistency. In dynamic sequences like \textit{car-roundabout} and \textit{pigs}, RVM maintains stable cluster assignments for moving objects, resisting the ``flickering'' artifacts observed in \textbf{VideoMAE} and \textbf{VideoMAE v2}. While \textbf{DINOv2} produces high-quality semantic segments, it lacks temporal awareness; RVM matches this semantic stability while explicitly modeling the temporal evolution of the instance.

\paragraph{Video Object Segmentation (DAVIS-2017).}
We visualize the quality of spatiotemporal feature correspondences through non-parametric label propagation on the DAVIS-2017 validation set. Using a context queue of 20 frames and $k=7$ nearest neighbors, RVM demonstrates robust object segmentation capabilities. By leveraging the recurrent memory, the model effectively propagates ground-truth masks from the initial frame to subsequent timesteps. The learned representations exhibit strong temporal stability, maintaining precise object boundaries even in the presence of fast motion and partial occlusions.

\paragraph{Human Pose Tracking (JHMDB).}
To assess fine-grained motion understanding, we evaluate keypoint tracking on the JHMDB dataset. We propagate human joint annotations using the same protocol as DAVIS ($k=7$). RVM captures the structural articulation of the human body, tracking individual keypoints (e.g., wrists, elbows, knees) with high precision. The model's recurrent mechanism ensures that feature trajectories remain consistent over time, minimizing drift and correctly re-associating keypoints after temporary self-occlusions characteristic of complex human actions.

\paragraph{Video Instance Parsing (VIP).}
We further challenge the model with the Video Instance Parsing (VIP) benchmark, which requires dense semantic part propagation. Unlike object-level segmentation, this task demands distinguishing between adjacent intra-object parts such as arms, legs, and hair. For this denser task, we increase the retrieval neighborhood to $k=10$. RVM successfully propagates these fine-grained semantic labels, resulting in temporally coherent part segmentations that respect the underlying human geometry better than frame-independent baselines.

\begin{figure}
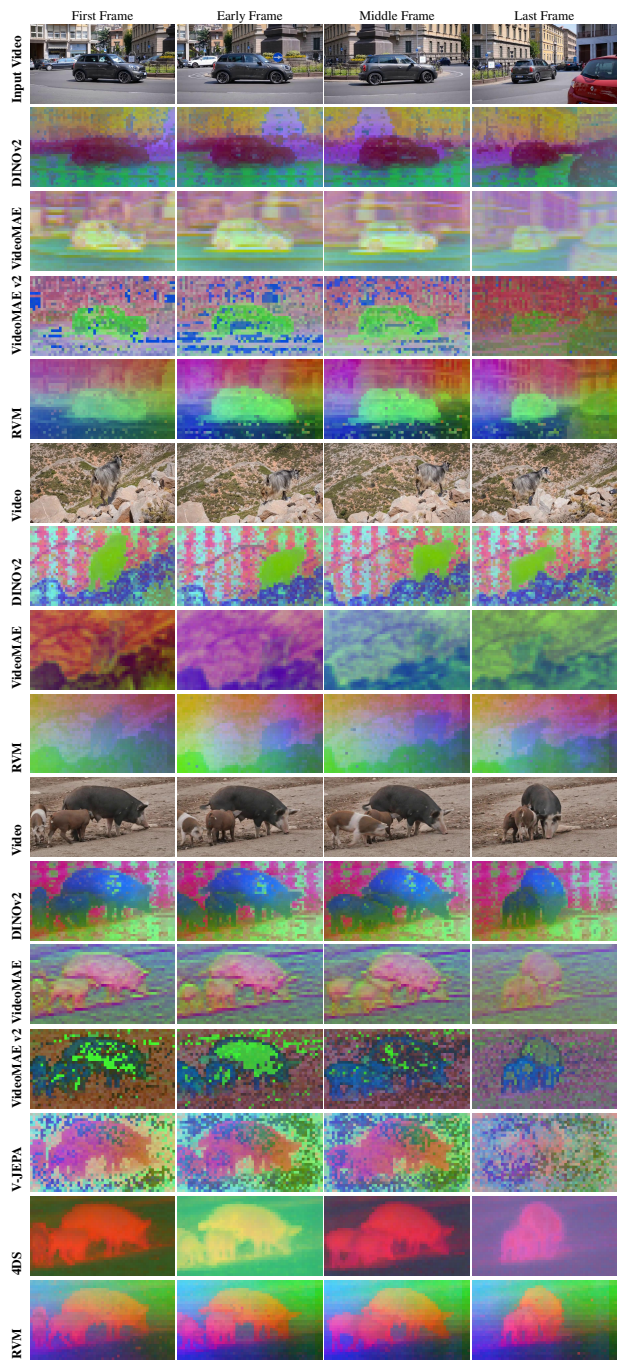

\centering
\newcommand{\imagerow}[7]{%
  \rotatebox{90}{\textbf{#1}} &
  \includegraphics[width=0.22\textwidth]{figures/davis/#2/#3_#4.pdf} &
  \includegraphics[width=0.22\textwidth]{figures/davis/#2/#3_#5.pdf} &
  \includegraphics[width=0.22\textwidth]{figures/davis/#2/#3_#6.pdf} &
  \includegraphics[width=0.22\textwidth]{figures/davis/#2/#3_#7.pdf} \\
}
\resizebox{0.99\linewidth}{!}{%
\begin{tabular}{l @{\hspace{0.5em}} c @{\hspace{0.2em}} c @{\hspace{0.2em}} c @{\hspace{0.2em}} c}
  & First Frame & Early Frame & Middle Frame & Last Frame \\

  \imagerow{Input Video}{frames}{car-roundabout}{0}{18}{37}{74}
  \imagerow{DINOv2}{dino2}{car-roundabout_pca}{0}{18}{37}{74}
  \imagerow{VideoMAE}{videomae}{car-roundabout_pca}{0}{18}{37}{74}
  \imagerow{VideoMAE v2}{videomae2}{car-roundabout_pca}{0}{18}{37}{74}

  \imagerow{RVM}{rvm}{car-roundabout_pca}{0}{18}{37}{74}  
  \imagerow{Video}{frames}{goat}{0}{22}{45}{89}
  \imagerow{DINOv2}{dino2}{goat_pca}{0}{22}{45}{89}
  \imagerow{VideoMAE}{videomae}{goat_pca}{0}{22}{45}{89}

  \imagerow{RVM}{rvm}{goat_pca}{0}{22}{45}{89}

  \imagerow{Video}{frames}{pigs}{0}{19}{39}{78}

  \imagerow{DINOv2}{dino2}{pigs_pca}{0}{19}{39}{78}

  \imagerow{VideoMAE}{videomae}{pigs_pca}{0}{19}{39}{78}
  \imagerow{VideoMAE v2}{videomae2}{pigs_pca}{0}{19}{39}{78}
  \imagerow{V-JEPA}{vjepa}{pigs_pca}{0}{19}{39}{78}

  \imagerow{4DS}{4ds}{pigs_pca}{0}{19}{39}{78}
  \imagerow{RVM}{rvm}{pigs_pca}{0}{19}{39}{78}  
\end{tabular}%
}
\caption{\textbf{Intrinsic Dimensionality and Smoothness.} We project the top-3 principal components of the frozen features to RGB space. \textbf{RVM} exhibits smooth color gradients that naturally follow the object's geometry, indicating a representation that is both spatially coherent and semantically meaningful. Conversely, features from other models often appear fragmented, lacking clear separation between the foreground and background.}
\label{fig:davis_pca}
\end{figure}

\begin{figure}
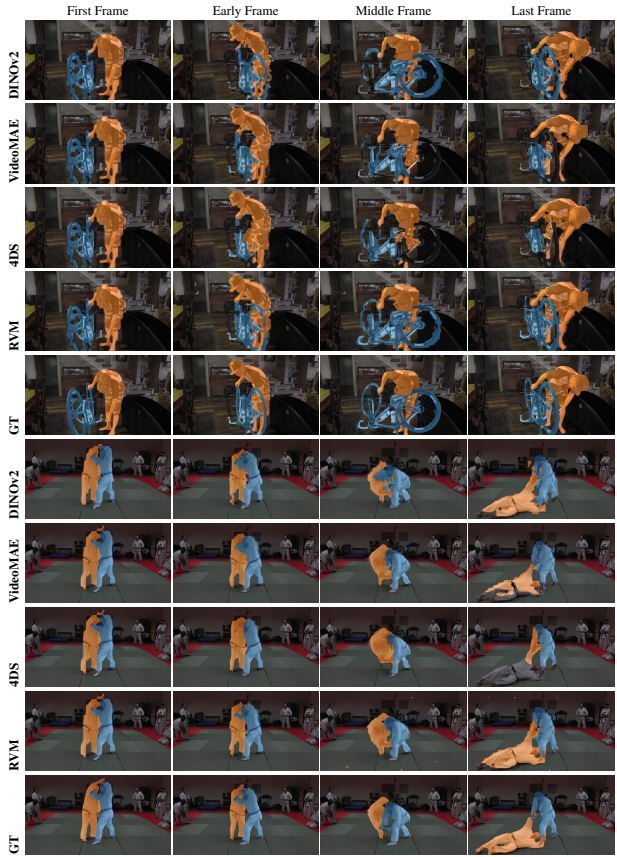

\centering
\newcommand{\imagerow}[7]{%
  \rotatebox{90}{\textbf{#1}} &
  \includegraphics[width=0.22\textwidth]{figures/davis_results/#2/#3_#4.pdf} &
  \includegraphics[width=0.22\textwidth]{figures/davis_results/#2/#3_#5.pdf} &
  \includegraphics[width=0.22\textwidth]{figures/davis_results/#2/#3_#6.pdf} &
  \includegraphics[width=0.22\textwidth]{figures/davis_results/#2/#3_#7.pdf} \\
}
\resizebox{0.99\linewidth}{!}{%
\begin{tabular}{l @{\hspace{0.5em}} c @{\hspace{0.2em}} c @{\hspace{0.2em}} c @{\hspace{0.2em}} c}
  & First Frame & Early Frame & Middle Frame & Last Frame \\
  \imagerow{DINOv2}{dino2}{bike-packing}{0}{17}{34}{68}
  \imagerow{VideoMAE}{videomae}{bike-packing}{0}{17}{34}{68}

  \imagerow{4DS}{4ds}{bike-packing}{0}{17}{34}{68}
  \imagerow{RVM}{rvm}{bike-packing}{0}{17}{34}{68}
  \imagerow{GT}{gt}{bike-packing}{0}{17}{34}{68}  
 
  \imagerow{DINOv2}{dino2}{judo}{0}{8}{17}{33}
  \imagerow{VideoMAE}{videomae}{judo}{0}{8}{17}{33}

  \imagerow{4DS}{4ds}{judo}{0}{8}{17}{33}
  \imagerow{RVM}{rvm}{judo}{0}{8}{17}{33}
  \imagerow{GT}{gt}{judo}{0}{8}{17}{33}

\end{tabular}%
}
\caption{\textbf{Qualitative evaluation on DAVIS-2017.} We propagate segmentation masks using nearest-neighbor retrieval ($k=7$) from a context queue of 20 frames. RVM (Ours) maintains accurate object boundaries and temporal consistency compared to baselines like VideoMAE and 4DS, which often exhibit mask degradation or flickering.}
\label{fig:davis_result}
\end{figure}

\begin{figure}
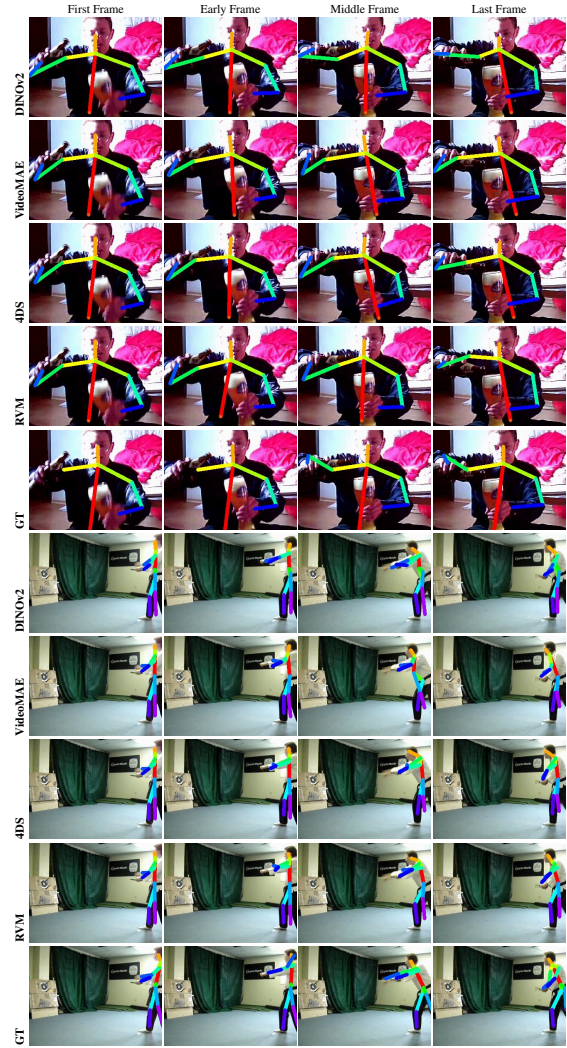

\centering
\newcommand{\imagerow}[7]{%
  \rotatebox{90}{\textbf{#1}} &
  \includegraphics[width=0.22\textwidth]{figures/jhmdb_results/#2/#3_#4.pdf} &
  \includegraphics[width=0.22\textwidth]{figures/jhmdb_results/#2/#3_#5.pdf} &
  \includegraphics[width=0.22\textwidth]{figures/jhmdb_results/#2/#3_#6.pdf} &
  \includegraphics[width=0.22\textwidth]{figures/jhmdb_results/#2/#3_#7.pdf} \\
}
\resizebox{0.9\linewidth}{!}{%
\begin{tabular}{l @{\hspace{0.5em}} c @{\hspace{0.2em}} c @{\hspace{0.2em}} c @{\hspace{0.2em}} c}
  & First Frame & Early Frame & Middle Frame & Last Frame \\

  \imagerow{DINOv2}{dino2}{Bier_richtig_einschenken_pour_u_cm_np1_fr_med_1}{0}{7}{15}{29}
  \imagerow{VideoMAE}{videomae}{Bier_richtig_einschenken_pour_u_cm_np1_fr_med_1}{0}{7}{15}{29}

  \imagerow{4DS}{4ds}{Bier_richtig_einschenken_pour_u_cm_np1_fr_med_1}{0}{7}{15}{29}
  \imagerow{RVM}{rvm}{Bier_richtig_einschenken_pour_u_cm_np1_fr_med_1}{0}{7}{15}{29}
  \imagerow{GT}{gt}{Bier_richtig_einschenken_pour_u_cm_np1_fr_med_1}{0}{7}{15}{29}
 
  \imagerow{DINOv2}{dino2}{KnifeThrowing_throw_f_nm_np1_le_med_2}{0}{9}{19}{38}
  \imagerow{VideoMAE}{videomae}{KnifeThrowing_throw_f_nm_np1_le_med_2}{0}{9}{19}{38}

  \imagerow{4DS}{4ds}{KnifeThrowing_throw_f_nm_np1_le_med_2}{0}{9}{19}{38}
  \imagerow{RVM}{rvm}{KnifeThrowing_throw_f_nm_np1_le_med_2}{0}{9}{19}{38}
  \imagerow{GT}{gt}{KnifeThrowing_throw_f_nm_np1_le_med_2}{0}{9}{19}{38}
\end{tabular}%
}
\caption{\textbf{Keypoint tracking on JHMDB.} The model propagates 15 human joint locations using label propagation ($k=7$, $\tau=0.7$). RVM accurately tracks rapid limb movements and maintains the structural consistency of the pose, distinguishing left/right limbs more effectively than baseline models like VideoMAE and 4DS.}
\label{fig:jhmdb_result}
\end{figure}

\begin{figure}
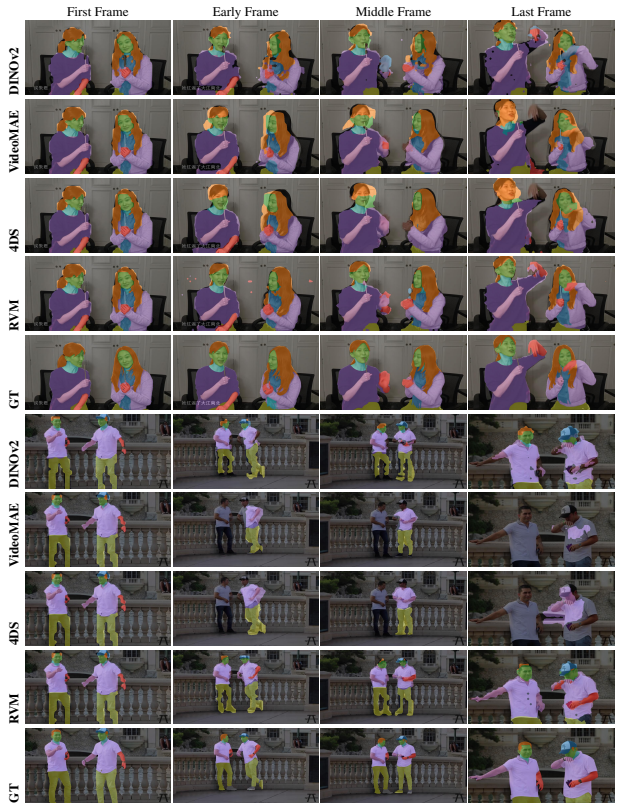

\centering
\newcommand{\imagerow}[7]{%
  \rotatebox{90}{\textbf{#1}} &
  \includegraphics[width=0.22\textwidth]{figures/vip_results/#2/#3_#4.pdf} &
  \includegraphics[width=0.22\textwidth]{figures/vip_results/#2/#3_#5.pdf} &
  \includegraphics[width=0.22\textwidth]{figures/vip_results/#2/#3_#6.pdf} &
  \includegraphics[width=0.22\textwidth]{figures/vip_results/#2/#3_#7.pdf} \\
}
\resizebox{0.99\linewidth}{!}{%
\begin{tabular}{l @{\hspace{0.5em}} c @{\hspace{0.2em}} c @{\hspace{0.2em}} c @{\hspace{0.2em}} c}
  & First Frame & Early Frame & Middle Frame & Last Frame \\
  \imagerow{DINOv2}{dino2}{videos361}{0}{5}{11}{22}
  \imagerow{VideoMAE}{videomae}{videos361}{0}{5}{11}{22}

  \imagerow{4DS}{4ds}{videos361}{0}{5}{11}{22}
  \imagerow{RVM}{rvm}{videos361}{0}{5}{11}{22}
  \imagerow{GT}{gt}{videos361}{0}{5}{11}{22}  
  \imagerow{DINOv2}{dino2}{videos233}{0}{7}{14}{28}
  \imagerow{VideoMAE}{videomae}{videos233}{0}{7}{14}{28}

  \imagerow{4DS}{4ds}{videos233}{0}{7}{14}{28}
  \imagerow{RVM}{rvm}{videos233}{0}{7}{14}{28}
  \imagerow{GT}{gt}{videos233}{0}{7}{14}{28}

\end{tabular}%
}
\caption{\textbf{Semantic part propagation on VIP.} We visualize the propagation of dense part labels (arm, leg, hair, etc.) using $k=10$ nearest neighbors. RVM distinguishes fine-grained semantic parts and tracks them consistently across the video clip, whereas other methods often confuse adjacent parts (e.g., arm vs. torso).}
\label{fig:vip_result}
\end{figure}


\end{document}